\documentclass[10pt]{article}
\usepackage[a4paper, total={6in, 9in}]{geometry}

\usepackage{amsmath,amsfonts}
\usepackage[utf8]{inputenc} 
\usepackage[T1]{fontenc}    
\usepackage{hyperref}       
\usepackage{booktabs}       
\usepackage{nicefrac}       
\usepackage{microtype}      
\usepackage{xcolor}         
\usepackage{amssymb}
\usepackage{mathtools}
\usepackage{amsthm}
\usepackage[capitalize,noabbrev]{cleveref}
\usepackage[sort&compress]{natbib}

\usepackage{caption}
\usepackage{subcaption}

\usepackage[inline]{enumitem}
\usepackage{xspace}
\newcommand{\eg}{\textit{e.g.}\@\xspace}
\newcommand{\ie}{\textit{i.e.}\@\xspace}
\newcommand{\iid}{i.i.d.\@\xspace}

\theoremstyle{plain}
\newtheorem{theorem}{Theorem}
\newtheorem{lemma}[theorem]{Lemma}

\newcommand{\R}{\mathbb{R}}

\newcommand{\EE}{\mathbb{E}}
\newcommand{\prob}{\mathbb{P}}

\newcommand{\var}{{\rm Var}}

\newcommand{\sign}{{\rm sgn}}

\newcommand{\eqdef}{\doteq}

\newcommand{\vect}[1]{\boldsymbol{\mathbf{#1}}}

\newcommand{\spa}{{\rm sp}}
\newcommand{\tim}{{\rm tm}}
\newcommand{\vx}{\vect x}
\newcommand{\vy}{\vect y}
\newcommand{\vr}{\vect r}
\newcommand{\vw}{\vect w}
\newcommand{\vg}{\vect g}
\newcommand{\vs}{\vect s}

\newcommand{\vm}{\vect m}

\newcommand{\mytitle}{Assessment of Spatio-Temporal Predictors in the Presence of Missing and Heterogeneous Data}

\date{}

\usepackage{authblk}
\author[1]{Daniele Zambon\thanks{Corresponding author, \texttt{daniele.zambon@usi.ch} .}}
\author[1,2]{Cesare Alippi}
\affil[1]{\small Universit\`a della Svizzera italiana, IDSIA, Switzerland.}
\affil[2]{\small Politecnico di Milano, Italy.}

\title{\bfseries\mytitle}

\begin{document}

\maketitle

\begin{abstract}
Deep learning methods achieve remarkable predictive performance in modeling complex, large-scale data. However, assessing the quality of derived models has become increasingly challenging, as more classical statistical assumptions may no longer apply. These difficulties are particularly pronounced for spatio-temporal data, which exhibit dependencies across both space and time and are often characterized by nonlinear dynamics, time variance, and missing observations, hence calling for new accuracy assessment methodologies. 
This paper introduces a residual correlation analysis framework for assessing the optimality of spatio-temporal relational-enabled neural predictive models, notably in settings with incomplete and heterogeneous data. By leveraging the principle that residual correlation indicates information not captured by the model, enabling the identification and localization of regions in space and time where predictive performance can be improved. A strength of the proposed approach is that it operates under minimal assumptions, allowing also for robust evaluation of deep learning models applied to multivariate time series, even in the presence of missing and heterogeneous data.
In detail, the methodology constructs tailored spatio-temporal graphs to encode sparse spatial and temporal dependencies and employs asymptotically distribution-free summary statistics to detect time intervals and spatial regions where the model underperforms. The effectiveness of what proposed is demonstrated through experiments on both synthetic and real-world datasets using state-of-the-art predictive models.
\end{abstract}

\begingroup
\renewcommand\thefootnote{}  
\footnotetext{%
~\\[0.1em]
This article is published in Neurocomputing (Elsevier), 2026, under the CC BY 4.0 license.
The final version of record is available at: \url{https://doi.org/10.1016/j.neucom.2026.132963}}
\endgroup

\section{Introduction}
\label{sec:intro}

Spatio-temporal predictive neural models fit data by leveraging the inductive biases inherent in possibly latent relational information derived from the temporal and spatial dimensions of observations~\citep{seo2018structured, bai2020adaptive, yu2018spatio, wu2019graph, pal2021rnn, ruiz2020gated, jin2023survey}. The spatial domain is often represented as a graph, capturing structures such as pixel grids, 3D meshes, road maps, or brain networks~\citep{shuman2013emerging, li2018diffusion, stankovic2020graph}. However, the term \emph{spatial} should be interpreted more broadly, encompassing functional dependencies among sensors that go beyond correlations tied to their physical positions~\citep{cini2023graph}.
Modern spatio-temporal data, such as multivariate time series generated by sensor networks, present significant challenges, including irregular sampling, substantial missing observations, and the time variant nature of heterogeneous sensors that can be added or removed over time~\citep{alippi2014intelligence, montero2021principles}.
In these contexts, predictive deep neural models may perform inconsistently across different regions of the spatio-temporal domain, complicating model quality assessments and the detection of unexpected behaviors. Assessing the optimality of these predictors, particularly when leveraging advanced processing neural architectures, remains an intricate task for which no robust and effective methods currently exist.

The quality of deep predictive models is typically assessed using task-specific accuracy metrics, with the squared error being a common choice, evaluating the 2-norm of the prediction \emph{residuals} -- the difference between the observed values and the model's predictions. Alternative metrics include absolute (MAE) and relative errors (MAPE)~\citep{gneiting2011making}. 
These approaches are practical and straightforward, likely contributing to their widespread adoption. However, they are inherently comparative, selecting the best model based on statistically superior performance relative to others. Consequently, they offer no direct insight into model optimality or specific guidance on areas needing improvement.

An alternative approach to assessing model quality focuses on analyzing the \emph{correlations} among prediction residuals rather than their magnitude. The rationale is that correlated residuals indicate structural information that the model has failed to capture~\citep{li2003diagnostic}, hence suggesting room for improvement.
Over the years, various hypothesis tests have been developed to detect residual dependencies~\citep{durbin1950testing, ljung1978measure, box2015time}. 
Commonly known as randomness or whiteness tests, these tests assess whether residuals exhibit white noise behavior, meaning they lack correlations. However, they rely on strict assumptions, including fully available multivariate time series, synchronous sampling, and identically distributed data.
In practice, however, incomplete and heterogeneous data are the norm, 
creating significant challenges for existing methods and underscoring the need for more robust tests~\citep{zambon2022aztest}.

\subsection{Contributions}
This paper presents a novel residual analysis framework for assessing the quality of models designed for spatio-temporal prediction tasks, particularly in scenarios with missing and heterogeneous data and dynamic (spatial) relational structures. 
The framework not only detects residual correlations but also pinpoints specific regions where models fail to capture the underlying data-generating process, offering a more nuanced understanding than traditional approaches.
In particular, the paper addresses three key questions about model optimality, intended here as the absence of autocorrelations and cross-dependencies in prediction residuals.
\begin{enumerate}[label=\textbf{Q\arabic*}]
\item \label{q:global} \emph{Is the nonlinear neural model optimal in terms of the absence of autocorrelations and cross-dependencies among prediction residuals?}
\item \label{q:nodes} \emph{Are there specific spatial regions (\eg, groups of time series) where predictions can be improved?}
\item \label{q:times} \emph{Are there specific time intervals where the model fails?} 
\end{enumerate}

The proposed residual analysis, referred to as AZ-analysis, addresses these questions by building on the statistic underlying the AZ-whiteness test~\citep{zambon2022aztest}. This enables the method to inherit the flexibility of the original test in handling complex spatio-temporal data, without requiring prior knowledge of the data distribution or identical distributions across time series. 
At the same time, AZ-analysis is conceptually distinct from the AZ-whiteness test, repurposing the underlying test statistic to construct interpretable, region-wise correlation measures.
In particular, AZ-analysis introduces tailored subgraphs that partition residuals into spatial and temporal regions and computes summary statistics, or \emph{scores}, to quantify local correlation and enable comparisons across different regions of the data. 
Building on these new elements, the paper develops a comprehensive framework for analyzing residual correlations at multiple levels, enabling the identification and localization of temporal drifts, node-specific dynamics not captured by the model, and anomalies rooted in data acquisition. The AZ-analysis also offers practical guidelines to interpret the results and address the identified correlation patterns.
The key contributions of this paper are as follows:
\begin{itemize}
    \item A method to identify heterogeneous time series exposing missing data whose associated prediction residuals exhibit significant evidence of correlation. 
    \item A method to pinpoint time intervals where residuals display correlation. 
    \item A method to identify spatio-temporal regions where residual correlations are particularly prominent.
\end{itemize}

The AZ-analysis is validated on synthetic data, and its practical relevance is demonstrated in real-world scenarios of traffic flow and energy production forecasting.

\subsection{Significance and impact}

Requiring minimal assumptions, the proposed approach is versatile and broadly applicable to real-world scenarios involving deep and graph-based predictors. 
Notably, it does not impose assumptions on the distribution of residuals or require data to be identically distributed. 
The only prerequisite is that residuals are centered at zero -- an assumption that is typically satisfied in most practical settings. This is one of the main advances with respect to the related literature, as further elaborated in Section~\ref{sec:related-work}.
Another key strength of the proposed analysis lies in its graph-based processing, which focuses on residual pairs that are more likely to exhibit correlations. This targeted approach enhances the statistical power of the method, making it more effective in detecting model shortcomings.

Although the proposed residual analysis does not quantify the magnitude of potential model improvements, it complements traditional accuracy-based evaluations. 
As demonstrated in Section~\ref{sec:experiment-traffic}, it provides an independent, metric-agnostic assessment of model quality. In particular, the experimental results reveal valuable insights from residual correlation analysis that standard prediction error evaluations fail to capture.

\medskip
The paper is structured as follows. Section~\ref{sec:related-work} reviews related work. Section~\ref{sec:preliminaries} introduces the edge scores, the foundation of the proposed residual analysis, and shows how to design the spatio-temporal graph -- a multiplex graph linking residual observations through spatial and temporal relationships.
Section~\ref{sec:az-test} provides an overview of the AZ-whiteness test.
The paper's novel contributions are detailed in Section~\ref{sec:az-analysis} and validated using synthetic data in Section~\ref{sec:synth}. 
Sections~\ref{sec:experiment-traffic} and \ref{sec:experiment-energy} showcase two real-world applications. Finally, conclusions are presented in Section~\ref{sec:conclusions}.

\section{Related work}
\label{sec:related-work}

Statistical tests designed to discover correlations in temporal data date back to Durbin and Watson's investigations~\citep{durbin1950testing,durbin1970testing}. Most of the literature has focused on univariate time series~\citep{geary1970relative,drouiche2000new}, although variants have appeared for inspecting longer-range correlations~\citep{box1970distribution,ljung1978measure} and multivariate time series~\citep{durbin1957testing,hosking1981equivalent,li2019testing}. These (classical) tests, however, often rely on strong assumptions, such as complete data, relatively low dimensionality, and identical distribution of observations, that may not hold in complex modern datasets.
More general statistical tests that leverage graph-based structures to connect data points can be considered. One of the earliest examples is the two-sample Friedman-Rafsky test~\citep{friedman1979multivariate}, which assesses the equality of distributions by looking at whether points sharing an edge belong to the same sample. 
Building on this foundation, subsequent research has extended these methods to change-point detection tests and statistical tests applicable to high-dimensional and non-Euclidean data~\citep{chen2013graphbased,chu2019asymptotic,chen2023graphbased}. 
By exploiting general data relationships, these graph-based tests remain effective in settings where traditional parametric tests break down, enabling hypothesis testing on complex data objects without requiring stringent assumptions about data dimensionality.

Orthogonal to the above, non-parametric tests using runs and sign statistics have long been used to check randomness in a sequence.
The \citet{wald1940test} runs test, and later the \citet{friedman1979multivariate} test, introduced the idea of counting how many adjacent observations come from the same sample. 
\citet{geary1970relative} employed sign statistics on prediction residuals as an indicator of serial correlation. More recently, \citet{zambon2022aztest} generalized the idea to spatio-temporal data, introducing a whiteness test to evaluate the overall presence of both spatial and temporal correlation in prediction residuals.

However, classical hypothesis tests provide only global summary statistics and do not localize the sources of correlation within space or time. With the recent rise of spatio-temporal prediction models in deep learning~\citep{jin2023survey,cini2023graph}, it has become necessary -- and is the focus of this paper -- to assess model quality at a finer-grained level, to better understand model behavior amid such complexity.
The works of \citet{papadimitriou2006local} and \citet{zhao2015inference} are two examples that have addressed local correlation patterns in the temporal domain. 
In contrast, the local indicators of spatial association (LISA) framework~\citep{anselin1995local} is a robust approach designed for spatial data, exploring local autocorrelation through statistics such as Moran's I~\citep{moran1950notes}, Geary's C~\citep{geary1954contiguity}, and multivariate ones~\citep{anselin2019local}. 
However, these approaches rarely consider scenarios involving missing or heterogeneous data distributions.

To the best of our knowledge, this paper is the first to provide a comprehensive spatio-temporal analysis of correlation patterns at the node (time series), temporal, and local spatio-temporal levels. 
In particular, the proposed AZ-analysis combines the local insight offered by the LISA framework with the flexibility of sign-based statistics, enabling wide applicability under mild assumptions.

\begin{figure}
\centering
\includegraphics[width=.6\columnwidth]{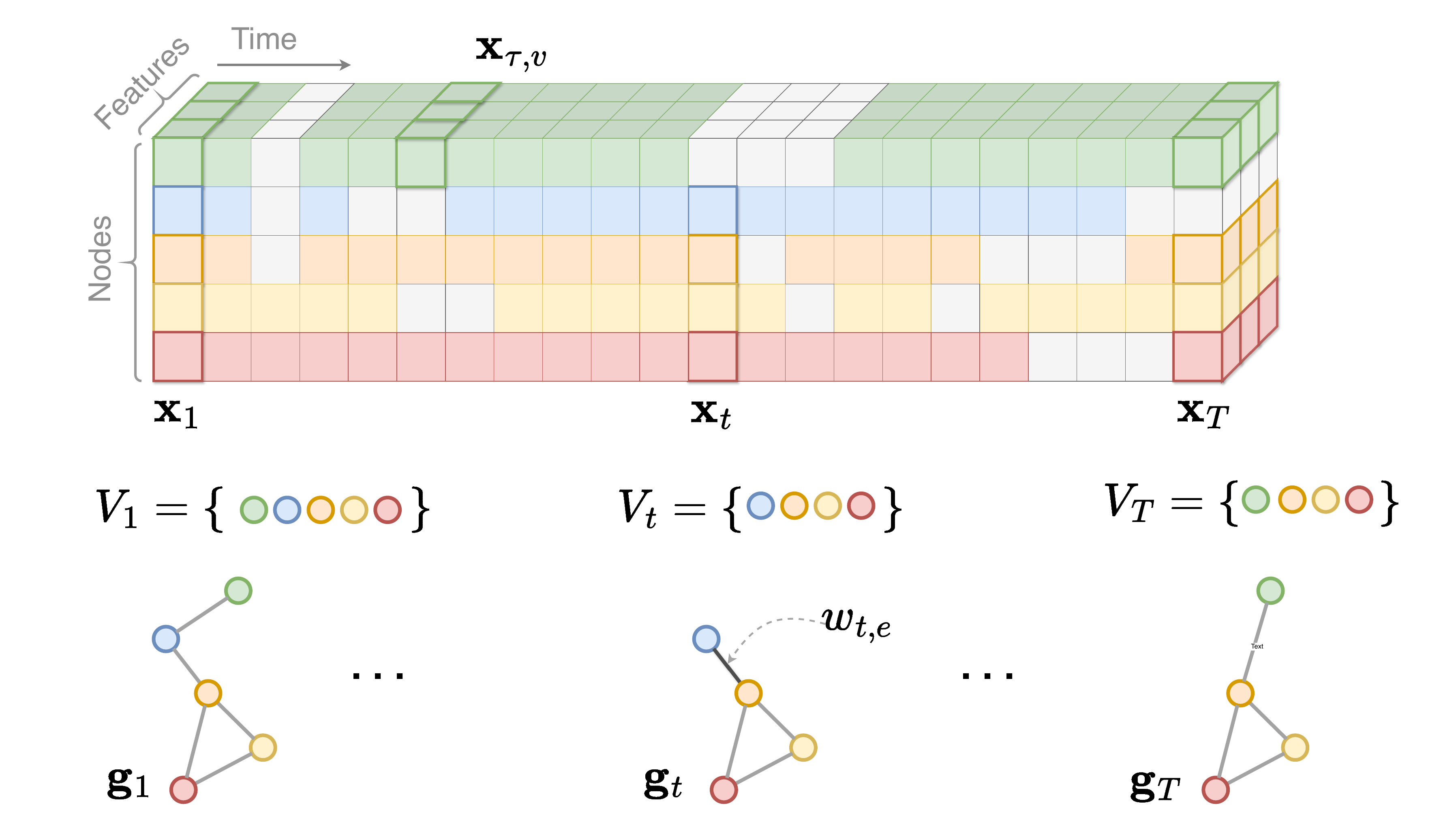}
    \caption{Representation of spatio-temporal data $\vx$ as a set of time series with associated sequence of graphs $(\vg_1,\vg_2,\dots,\vg_T)$ encoding functional relations. 
    Observation $\vx_{\tau,v}$ at time step $\tau$ and node/sensor $v$ is multivariate. Nodes need not be available at all times (light gray boxes) and the graph topology can vary. $w_{t,e}$ denotes the weight of edge $e$ at time step $t$. 
    }
    \label{fig:graph-data-small}
\end{figure}

\section{Preliminaries}
\label{sec:preliminaries}

\subsection{Spatio-temporal predictions}

Consider a multivariate time series 
\begin{equation}
\vx\eqdef\{\vx_{t,v}:\ v\in V_t,\;t=1,2,\dots\}
\end{equation}
where $\vx_{t,v}\in\R^{d_x}$ is a stochastic vector representing the observation of the $v$-th time series at time step $t$. 
We may refer to $v\in V_t$ as the multi-channel sensor associated with the $v$-th time series from the sensor set $V_t$ available at time step $t$; denote $\vx_t\eqdef\{\vx_{t,v}:v\in V_t\}$ and $V\eqdef\bigcup_t V_t$. 
Consider a family of predictive models $f_{\theta}$ with parameter vector $\theta$ trained to predict, at every time step $t$, target $\vy_{t,v}\in\R^{d_y}$ for all $v$. The inputs to these models come from $\vx$. Exogenous variables, such as those describing different types of sensors, are assumed to be encoded in $\vx$.
A common example of a predictive task is time series forecasting~\citep{cini2023graph}: in this context, $\vy_{t}$~corresponds to the window of subsequent observations $\vx_{t:t+H}$, therefore $d_y=d_x\times H$; notation $\vx_{a:b}$ indicates $\{\vx_{t}:a\le t< b\}$. 
The presented formulation naturally handles missing data in $\vx$ using the sets $V_t$, which may not necessarily be equal to $V$. 
Moreover, no specific homogeneity assumptions are imposed on the data, and the ability to handle heterogeneous sensors will be evident from the discussion in Section~\ref{sec:az-test}.

In this paper, we consider the existence of functional dependencies among observations either available or extracted from the data itself~\citep{cini2023sparse}. We encode these dependencies at time $t$ in graph 
\begin{equation}
\vg_t\eqdef(V_t,E_t,\vw_t, \vx_t)
\end{equation}
whose edges $e\in E_t\subseteq V_t\times V_t$ represent (directed or undirected) binary relations between nodes in $V_t$ at time $t$.
Each node $v\in V_t$ corresponds to time series observation $\vx_{t,v}\in\vx_t$.
Nonnegative scalar weights $w_{t,e}\in\vw_t$ might be associated with each edge $e\in E_t$ to encode the strength of the relation, \eg, the physical distance of the two sensors or the capacity of a link in a transportation network.
The result is a graph sequence
\begin{equation}\label{eq:graph-sequence}
(\vg_1,\vg_2,\dots,\vg_t,\dots),
\end{equation}
where node set $V_t$, edge topology $E_t$ and edge weights $\vw_t$ can change over time. 
A visual representation of the set of time series $\vx$ and their relations in $\{\vg_t\}$ is provided in Figure~\ref{fig:graph-data-small}.

Model predictions $\hat{\vy}_t = f_{\hat{\theta}}(\vx_{t-W:t})$ are typically obtained from a sliding window of input data, defined as  $\vx_{t-W:t}$, encompassing the past $W$ time steps. The parameter vector $\hat{\theta}$ is learned by minimizing a prediction loss function, such as mean squared error (MSE) or mean absolute error (MAE), between the predicted values $\hat{\vy}_{t,v}$ and the ground truth targets $\vy_{t,v}$.  
More broadly, predictions at time step $t$ can incorporate topological information ($\hat{\vy}_t = f_{\hat{\theta}}(\vg_{t-W:t})$) as in graph-based models like spatio-temporal graph neural networks (STGNNs)~\citep{cini2023graph}. Predictions may also depend on past predictions $\hat{\vy}_{t-W:t}$ and previous target values $\vy_{t-W:t}$.
Define prediction residuals as 
\begin{equation}
\vr_{t,v} \eqdef \vy_{t,v} - \hat\vy_{t,v} \ \in \mathbb R^{d_y}
\end{equation}
for all $v$ and $t$. The $i$-th component, $\vr_{t,v}(i)$, of vector $\vr_{t,v}$ is the residual associated with the $i$-th target variable $\vy_{t,v}(i)$.

\subsection{Spatio-temporal graphs}
\label{sec:multiplex}

A finite sequence $(\vg_1,\vg_2,\dots,\vg_T)$ of length $T$ and the associated set of residuals 
\begin{equation}\label{eq:residuals}
    \vr\eqdef\{\vr_{t,v} : v\in V_t, t=1,\dots,T\}
\end{equation}
can be represented as a multiplex graph 
\begin{equation}\label{eq:multiplex}
\vg^*\eqdef(V^*,E^*,\vw,\vr)
\end{equation}
constructed by stacking all graphs $\vg_t$, for all $t=1,2,\dots,T$. Each node at time $t$  is connected to itself at time $t-1$, as shown in Figure~\ref{fig:sign-dot}. In the figure, solid gray lines denote edges in $\vg_t$, while dashed gray lines indicate temporal edges across consecutive graphs. 
Moreover, $\vg^*$ is augmented with the residuals $\vr$ as node signals and a set $\vw$ of edge weights.
As we will see later, the spatio-temporal graph $\vg^*$ offers a convenient static representation of the observed system dynamics.

Formally, node set $V^*$ is defined as
\begin{equation}\label{eq:gstar-node-set}
    V^* \eqdef \left\{v_t \eqdef (t,v): v\in V_t,\ t=1,2,\dots,T\right\},
\end{equation}
where each node $v_t$ is associated with the residual vector $\vr_{t,v} \in\vr$.
Edge set $E^*$ is the union of sets
\begin{equation}
E^*\eqdef E_\spa \cup E_\tim
\end{equation}
where
\begin{align}
\label{eq:Esp-def}
E_\spa &\eqdef \left\{\{u_t, v_t\} : (u,v) \in E_t,v\ne u,1\le t\le T\right\},
\\
E_\tim &\eqdef \left\{\{v_t, v_{t+1}\} : v\in V_{t}\cap V_{t+1},1\le t < T\right\}.
\end{align}
$E_\spa$ collects all (spatial) edges in sets $E_t$, for all $t$, regardless of their orientation whenever it applies, while $E_\tim$ is the set of edges connecting corresponding nodes along the temporal dimension.
Note that edge directions and self-loops are excluded in \eqref{eq:Esp-def}. However, this has no impact on the following analysis, 
as the statistics derived from equation \eqref{eq:edge-sign} disregard directional information. Secondly, our focus is on the correlations between different observations in both time and space.
Finally, if $\vg_t$ is undirected, weight $w_{u_t,v_t}$ of spatial edge $\{u_t,v_t\}\in E_\spa$ in $\vg^*$  is set equal to weight $w_{t,\{u,v\}}$ of $\vg_t$, \ie, the weight corresponding to edge $\{u,v\}\in E_t$. 
Conversely, for a directed graph $\vg_t$, the weight $w_{u_t,v_t}$ equals $w_{t, (u,v)}$ or $w_{t, (v,u)}$, depending on which directional edge is in $E_t$. If both $(u,v)$ and $(v,u)$ are present, then $w_{u_t,v_t}=w_{t,(u,v)} + w_{t,(v,u)}$. 
The weights for temporal edges can be defined arbitrarily or, \eg, to balance the overall impact of the spatial and temporal edges, as discussed in Section~\ref{sec:az-test}. 

For clarity of presentation, Table~\ref{tab:notation} summarizes the main notation adopted throughout the paper.

\begin{table}
\caption{Summary of notation used in the paper.}
\label{tab:notation}
\centering
\small
\renewcommand{\arraystretch}{1.15}
\begin{tabular}{cl}
\hline
\textbf{Symbol} & \textbf{Description} \\
\hline
$\vx_{t,v} \in \mathbb{R}^{d_x}$ & Observation of sensor/node $v$ at time $t$ \\
$\vy_{t,v},\ \hat \vy_{t,v} \in \mathbb{R}^{d_y}$ & Target and model prediction at $(t,v)$ \\
$\vr_{t,v} \eqdef \vy_{t,v} - \hat \vy_{t,v}$ & Prediction residual at $(t,v)$ \\
$\vr \eqdef \{\vr_{t,v}:\forall v,t\}$ & Collection of all prediction residuals\\[2mm]

$\vg_t,\  V_t,\  E_t$ & Graph at time $t$ and corresponding node and edge sets\\
$\vw_t,\ w_{t,e}$ & Set of edge weights of $\vg_t$ and weight of edge $e \in E_t$ \\
$\vg^* \eqdef (V^*, E^*, \vw, \vr)$ & Multiplex graph with $\vr$ as node attributes\\
$E_{\mathrm{sp}},\  E_{\mathrm{tm}}$ & Spatial and temporal edge sets of $\vg^*$\\
$\vs=(V_s,E_s,\vw_s,\vr_s)$ & Generic subgraph of $\vg^*$ \\
$|E_s|,\ \lVert\vw_s\rVert_1,\ \lVert\vw_s\rVert_2$ & Number of edges, 1- and 2-norm of the edge weights of $\vs$\\[2mm]

$C_\lambda(\vs),\ c_\lambda(\vs)$ & AZ-whiteness test statistic and correlation score on $\vs$\\
$c_\lambda(v),\  c_\lambda(t),\  c_\lambda(v,t)$ & Node, time, and local scores at node $v$ and time $t$\\
$\lambda \in [0,1]$ & Parameter trading off spatial and temporal contributions\\
$\gamma>0,\  \alpha \in (0,1)$ & Threshold and significance level of the AZ-whiteness test\\
\hline
\end{tabular}
\end{table}

\section{AZ-whiteness test}
\label{sec:az-test}

The analysis presented in this paper leverages the AZ-whiteness test~\citep{zambon2022aztest}, a statistical test designed to detect the presence of both autocorrelation and cross-correlation in data streams.
The test is defined over hypotheses
\begin{equation}
\label{eq:test-H0-H1}
\begin{cases}
H_0:& 
\text{All pairs }(\vr_{t,v},\vr_{\tau, u}) \text{ are uncorrelated,}
\\
H_1:& 
\text{At least one pair }(\vr_{t,v},\vr_{\tau, u}) \text{  is correlated,}
\end{cases}
\end{equation}
for $(t,v)\ne (\tau, u)$, and rejects the null hypothesis $H_0$ whenever the absolute value of the statistic $C_\lambda(\vg^*)$ exceeds a threshold $\gamma$. This threshold is predetermined according to a user-defined significance level $\alpha$
\begin{equation}\label{eq:significance-level}
\prob\left(\,|C_\lambda(\vg^*)|\ge\gamma\mid H_0\,\right)=\alpha.
\end{equation}

The AZ-whiteness test statistic $C_\lambda(\vg^*)\in\R$ is computed on the prediction residuals $\{\vr_{t,v}:\forall\ t,v\}$ with reference to the spatio-temporal multiplex graph $\vg^*$.
In more detail, the test statistic is defined as
\begin{equation}\label{eq:az-test-statistic}
C_\lambda(\vg^*) \eqdef 
\frac{
    \lambda \widetilde C_\spa + (1- \lambda)\widetilde C_\tim
}{
    \left( \lambda^2 W_{\spa} + (1- \lambda)^2W_{\tim}\right)^{1/2}
},
\end{equation}
where parameter $\lambda\in [0,1]$ trades off spatial and temporal contributions
\begin{align}\label{eq:C-tilde-space}
\widetilde C_\spa &\eqdef\sum_{\{u_t,v_t\}\in E_\spa} w_{\{u_t,v_t\}}\; \sign\left(\vr_{t,u}^\top \vr_{t,v}\right),
\\
\widetilde C_\tim &\eqdef\sum_{\{v_t,v_{t+1}\}\in E_\tim} w_\tim\; \sign\left(\vr_{t,v}^\top \vr_{t+1,v}\right),
\end{align}
respectively. 
Scalar quantities $\widetilde C_\spa$ and $\widetilde C_\tim$ are weighted sums assessing spatial and temporal correlation, respectively, by accounting for positive and negative signs
\begin{equation}\label{eq:edge-sign}
\sign\left(\vr_{t,v}^\top\vr_{\tau,u}\right)\ \in \{-1,0,1\}
\end{equation}
of the scalar product 
\begin{equation}
\vr_{t,v}^\top\vr_{\tau,u}=\sum_{i=1}^{d_y} r_{t,v}(i)\;r_{\tau,u}(i)
\end{equation}
between residual vectors $\vr_{t,v}$ and $\vr_{\tau,u}$; sign function
$\sign(a)$ equals $-1, 0$ or $1$ depending on whether $a$ is negative, null, or positive. 
Scalar $w_\tim$ is a single weight assigned to all temporal edges and set such that the normalization term
$W_\tim\eqdef|E_\tim|\cdot w_\tim^2$ equals $W_\spa$  
\begin{equation}\label{eq:W-space}
W_{\spa} \eqdef  \sum_{\{u_t,v_t\}\in E_\spa} w_{\{u_t,v_t\}}^2.
\end{equation}

\begin{figure}
    \centering
    \includegraphics[width=.54\columnwidth]{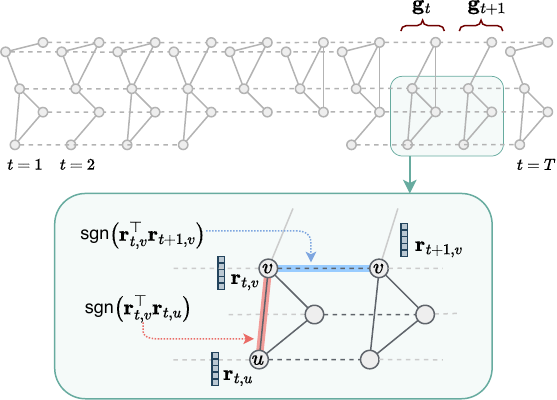}
    \caption{A section view of the spatio-temporal graph $\vg^*$ from Section~\ref{sec:multiplex}.
    Each node is associated with a residual vector $\vr_{t,v}$ and each edge -- either 
    spatial (red) or temporal (blue) -- is associated with a sign~\eqref{eq:edge-sign}.}
    \label{fig:sign-dot}
\end{figure}

\begin{figure}
    \centering
    \begin{subfigure}[b]{.49\textwidth}
    \includegraphics[width=\textwidth]{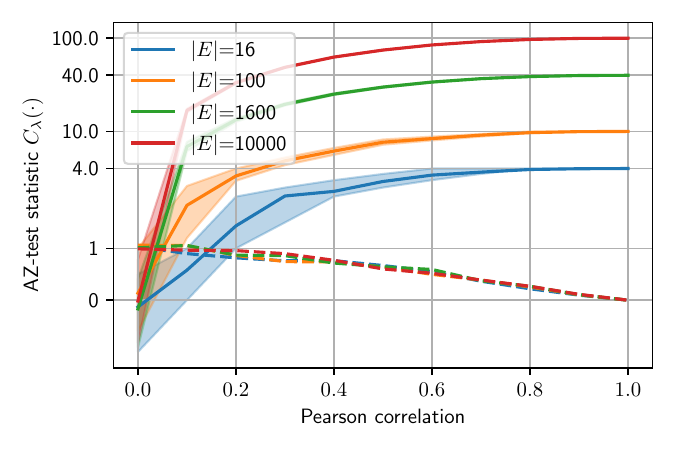}
    \caption{}
    \label{fig:score-growth-C}
    \end{subfigure}
    \begin{subfigure}[b]{.49\textwidth}
    \includegraphics[width=\textwidth]{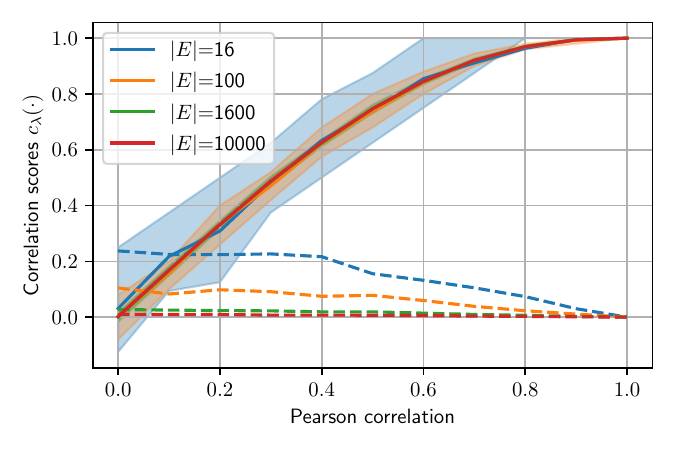}
    \caption{}
    \label{fig:score-growth-c}
    \end{subfigure}
    \caption{The figure compares the value of statistics $C_\lambda(\cdot)$ defined in \eqref{eq:az-test-statistic} (left panel, \ref{fig:score-growth-C}) with that of the correlation scores $c_\lambda(\cdot)$ of \eqref{eq:az-max-adjusted} (right panel, \ref{fig:score-growth-c}). Different levels of residual correlation and the number $|E|$ of graph edges are considered.
    Each color corresponds to a different number of edges. Solid lines represent the expected value of the score estimated over 100 repeated experiments, dashed lines the standard deviation, shaded area the interquartile interval.}
    \label{fig:score-growth}
\end{figure}

The distinction between spatial and temporal contributions is also highlighted in Figure~\ref{fig:sign-dot}.
The intuition behind the test statistic \eqref{eq:az-test-statistic} is that large positive values of $C_\lambda(\vg^*)$ indicate similarly oriented adjacent residuals, suggesting positive correlation among variables (see also Figure~\ref{fig:score-growth}). Likewise, strongly negative values of $C_\lambda(\vg^*)\ll 0$ indicate negative correlation.

Identifying a value of $\gamma$ for any choice of $\alpha$ is possible thanks to Theorem~\ref{theo:az-test}. This theorem shows that, asymptotically with the number of edges, $C_\lambda(\vg^*)$ follows a standard Gaussian distribution. Importantly, this result holds irrespective of the distribution of the residuals 
and does not require them to be identically distributed. 
\begin{theorem}[\citet{zambon2022aztest}]\label{theo:az-test}
Consider a spatio-temporal graph $\vg^*$ with associated stochastic residuals $\vr$ and the hyperparameter $\lambda\in[0,1]$. Assume
\begin{enumerate}[label=\textbf{A\arabic*}, topsep=0pt, parsep=0pt, itemsep=0pt] 
    \item \label{a:indep} All residual vectors $\vr_{t,v}$ in $\vr$ to be mutually independent and almost surely $\neq \vect 0$,
    \item \label{a:median} $\EE_{\vr_{t,v}}\left[\sign\left(\bar\vr^\top\vr_{t,v}\right)\right] = 0$ for all $\bar \vr\in\mathbb R^{d_y}\setminus\{\vect 0\}$ and $v_t \in V^*$,
    \item \label{a:weights} $w_\tim,w_{t,\{u,v\}}\in [w_-,w_+]$ for all $\{u_t,v_t\}\in E_t$ and $t$, with $0<w_-<w_+$, 
    \end{enumerate}
then, the distribution of $C_\lambda(\vg^*)$ in \eqref{eq:az-test-statistic} converges weakly to a standard Gaussian distribution $\mathcal N(0, 1)$ as the
number $|E^*|$ of edges goes to infinity.
\end{theorem}
The proof exploits the central limit theorem under the Lindeberg condition and is detailed in the original paper~\citep{zambon2022aztest}.

The assumptions of Theorem~\ref{theo:az-test} are very mild. This is partly due to the choice of assessing correlation using signs, as defined in~\eqref{eq:edge-sign}.
More specifically, assumption \ref{a:indep} is required to satisfy the null hypothesis.
In contrast, \ref{a:median} and \ref{a:weights} are straightforwardly resolved in most scenarios, if not already fulfilled~\citep{zambon2022aztest}.
In particular, for scalar residuals, \ref{a:median} reduces to requesting that the residual median is zero. 
Lastly, \ref{a:weights} requests weights to be positive and bounded, as they are assumed to encode the strength of the dependency between nodes. Whenever no weights are given, or if they do not satisfy \ref{a:weights}, we suggest considering setting all weights to $1$ or monotonically transforming them.

\section{AZ-analysis of residuals}
\label{sec:az-analysis}

The AZ-analysis introduced in this paper addresses questions \ref{q:global}--\ref{q:times} by inspecting families $\{\vs_i\}$ of subsets of all possible pairs of residuals in $\vr$. The goal is to identify those subsets that display stronger evidence of correlation, as measured by a scoring function $c_\lambda(\cdot)$, which is related to $C_\lambda(\cdot)$ as explained in the next subsection.
It is convenient to refer subsets $\vs_i$ as subgraphs of the fully connected graph over all observations in $\vr$.
Given suitably designed subgraphs, we compute and compare correlation scores $c_\lambda(\vs_i)$ for different subgraphs $\vs_i$, having in mind that $|c_\lambda(\vs_i)|>|c_\lambda(\vs_j)|$ implies the existence of a larger residual correlation in $\vs_i$ than in $\vs_j$. 
Although subgraphs $\{\vs_i\}$ may differ in size, we show that the associated scores $\{c_\lambda(\vs_i)\}$ can be compared and ranked.

The remainder of this section is structured as follows. We introduce the notion of correlation scores and explain how they enable comparisons across different residual subsets in Section~\ref{sec:def-corr-scores}. We then discuss the practical challenge of handling the large number of possible subgraphs and describe strategies for defining meaningful subgraph families in Section~\ref{sec:subgraph-number}. Subsequent subsections (Sections~\ref{sec:presence}--\ref{sec:spacetime-scores}) address the three key questions \ref{q:global}--\ref{q:times} related to the identification of residual correlations. Finally, we offer practical considerations and implementation guidance in Section~\ref{sec:practical-considerations}.

\subsection{Correlation scores}
\label{sec:comparability}\label{sec:def-corr-scores}

Under the null hypothesis, test statistic $C_\lambda(\vs)$ asymptotically follows a standard Gaussian distribution, regardless of the size of the graph $\vs$; see Theorem~\ref{theo:az-test}. 
This implies that all sample means should be centered around zero, except for those indicating evidence of correlation. Additionally, the test maintains the same standard deviation, ensuring a certain degree of comparability.

The left panel of Figure~\ref{fig:score-growth} illustrates values of $C_\lambda(\vs)$ as a function of the total number of edges $|E_{\vs}|$ in $\vs$ and the Pearson correlation between residuals. A detailed description of the experiment is provided in~\ref{a:score-growth}.
The figure illustrates that when no correlation exists among residuals, all scores are centered around zero with a standard deviation of 1, regardless of the number of edges considered. As correlation increases, each curve -- corresponding to a different number of edges -- exhibits a monotonic increasing trend.
However, we also observe that scores based on a larger number of edges are more effective in identifying correlated residuals. 
This suggests introducing a statistic that enables the comparison of score values independent of the number of edges used in their computation.

The maximum value of each curve in Figure~\ref{fig:score-growth} is $\sqrt{|E_{\vs}|}$. In fact, in the setting of the experiment, all weights (accounting for the factors $\lambda$ and $1-\lambda$ too) and all signs in \eqref{eq:az-test-statistic} are equal to 1. Therefore, the test statistic simplifies to
\begin{equation}
    C_\lambda(\vs) = \frac{\sum_{e\in E_\spa} 1 + \sum_{e\in E_\tim} 1}{\sqrt{\sum_{e\in E_\spa}1^2 +\sum_{e\in E_\tim}1^2}} = \frac{|E_{\vs}|}{\sqrt{|E_{\vs}|}}.
\end{equation}
More generally, the following lemma holds.
\begin{lemma}\label{lem:az-stat-max-min}
Denote $\vw_{\vs,\lambda}=\{\lambda w_e:e\in E_{\spa}\}\cup\{(1-\lambda) w_e:e\in E_{\tim}\}$ the set of all spatial and temporal weights of $\vs$, adjusted according to $\lambda$.
The maximum and minimum values of $C_\lambda(\vs)$ are $\pm \lVert\vw_{\vs,\lambda}\rVert_1/\lVert\vw_{\vs,\lambda}\rVert_2$.
\end{lemma}
\begin{proof}
The numerator and denominator in \eqref{eq:az-test-statistic} satisfy
\begin{align*}
    \left|\lambda\widetilde{C}_\spa +(1-\lambda)\widetilde{C}_\tim \right| &\le   \sum_{e\in E_{\spa}} |\lambda w_e| + \sum_{e\in E_{\tim}} |(1-\lambda) w_e|  = \lVert\vw_{\vs,\lambda}\rVert_1
    \\
    \lambda^2W_{\spa} +(1-\lambda)^2 W_{\tim} &= \lVert\vw_{\vs,\lambda}\rVert_2^2.
\end{align*}
In particular, the bounds are tight and are achieved when the signs are either all positive ($+1$) or all negative ($-1$). Therefore,  $|C_\lambda(\vs)| \le {\lVert\vw_{\vs,\lambda}\rVert_1}/{\lVert\vw_{\vs,\lambda}\rVert_2}$.
\end{proof}

Based on Lemma~\ref{lem:az-stat-max-min} above, we scale the statistic $C_\lambda(\cdot)$ so that its values fall within the interval $[-1, 1]$ as
\begin{equation}\label{eq:az-max-adjusted}
    c_\lambda(\vs) \eqdef C_\lambda(\vs) \frac{\lVert \vw_{\vs,\lambda}\rVert_2}{\lVert \vw_{\vs,\lambda}\rVert_1} 
    =  
\frac{
    \lambda \widetilde C_\spa + (1- \lambda)\widetilde C_\tim
}{
    \lambda \sum_{e\in E_{\spa}} w_e + (1- \lambda) \sum_{e\in E_{\tim}} w_e
};
\end{equation}
call $c_\lambda(\cdot)$ the \emph{correlation scores}.

The right panel of Figure~\ref{fig:score-growth} shows the behavior of scores $c_\lambda(\cdot)$. 
While scores share the same advantages as $C_\lambda(\cdot)$, in that their magnitude increases with the residual correlation, we also observe that their expected values do not depend on the number of edges. 
In particular, $\mathbb E[c_\lambda(\vs)]=0$ under Assumptions \ref{a:indep}~and~\ref{a:median}.
Accordingly, we suggest inspecting those subgraphs $\vs_i$ associated with $|C_\lambda(\vs_i)|\gg 0$ (evidencing correlation, as discussed in Section~\ref{sec:az-test}), 
starting from those with large $|c_\lambda(\vs_i)|$ in decreasing order; indeed, checking the presence of correlation first avoids inspecting and comparing scores whose values do not reflect any significant correlation.

\subsection{Subgraphs as sets of edges}\label{sec:subgraph-number}
Note that the number of potentially correlated pairs of residuals is, in principle,
quadratic in the number of residuals $\vr_{t,v}$ in \eqref{eq:residuals}. Denoting by $T$ the length of the time series and by $N$ the average number of available sensors, there are $N\, T$ observed residuals, which leads to $(N\,T)(N\,T-1)/2$ pairs.
Indeed, considering all possible subsets of node pairs is, in practice, infeasible, since their number is approximately $2^{{N^2\,T^2/2}}$. We address this complexity problem in two steps.

First, we confine the analysis to node pairs corresponding to edges in $\vg^*$,
which amounts to considering all possible subgraphs of $\vg^*$. 
The relational information encoded in multiplex $\vg^*$ is relevant for solving the downstream task. Therefore, the node pairs represented by its edges are more likely to exhibit correlation in the associated residuals. The inspection reduces to $2^{|E_*|}$ subgraphs, \ie, approximately $2^{N\,\delta\,T / 2}$, with $\delta$ the average node degree in the graph sequence incremented by 2 to account for temporal edges. 
Second, 
even analyzing all $2^{E_*}$ subsets remains impractical. Therefore, we propose principled families 
of subgraphs, each designed to address the three key questions \ref{q:global}, \ref{q:nodes}, and \ref{q:times}  
in Sections~\ref{sec:presence}--\ref{sec:spacetime-scores}.

\begin{figure}
\centering
\includegraphics[width=.7\textwidth]{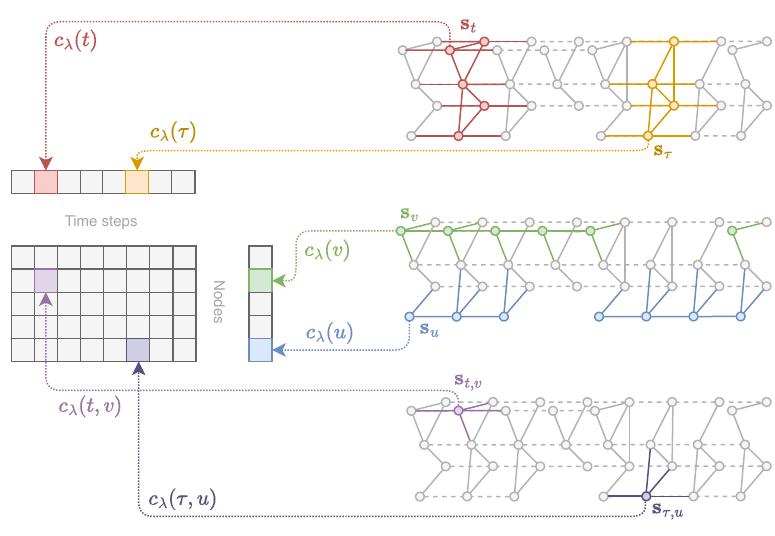}
    \caption{
    Top right) Generic subgraphs $\vs_t,\vs_\tau$ involved in the computation of time scores $c_\lambda(t)$ and $c_\lambda(\tau)$.
    Center right) Generic subgraphs $\vs_v,\vs_u$ involved in the computation of node scores $c_\lambda(v)$ and $c_\lambda(u)$.
    Bottom right) Generic subgraphs $\vs_{t,v},\vs_{\tau,u}$ involved in the computation of local spatio-temporal scores $c_\lambda(v,t)$ and $c_\lambda(\tau,u)$.
    Left) Convenient arrangement of the scores for visualization purposes; see subsequent experimental sections.}
    \label{fig:subset-spacetime}
    \label{fig:subset-space}
    \label{fig:subset-time}
\end{figure}

\subsection{Determining the presence of correlation (\ref{q:global})}\label{sec:presence}

The AZ-whiteness test \eqref{eq:az-test-statistic} with $\lambda=1/2$ applied to multiplex graph $\vg^*$ allows us to assess the overall presence of correlation; here, large values of $|C_{\lambda=1/2}(\vg^*)|$ denote a pronounced correlation in the residuals.
We select the value $\lambda=1/2$ as the default, since it weighs the spatial and temporal components equally; see the definition in \eqref{eq:az-test-statistic}.
Instead, the values of the statistic $C_{\lambda}(\vg^*)$ with $\lambda=0$ and $\lambda=1$ provide first insights about the presence of temporal and spatial correlations, respectively. This follows from Theorem~\ref{theo:az-test}, as $C_{\lambda=0}(\vg^*)$ and $C_{\lambda=1}(\vg^*)$ share the same distribution under the assumption of independent residuals; therefore, comparing these values reveals whether temporal or spatial components in the processing pipeline have greater room for improvement. 

For instance, $|C_{\lambda=0}(\vg^*)| > |C_{\lambda=1}(\vg^*)|$ with $|C_{\lambda=0}(\vg^*)|\gg 0$ would indicate that temporal (auto)correlation is more problematic than spatial correlation. 
Should this be the case, the designer may intervene on the temporal processing pipeline to improve the model, \eg, by changing the width of the input window or the neural prediction architecture; we refer to~\citep{gao2022equivalence,cini2023graph} for discussions about different ways of combining spatial and temporal processing in deep architectures.

\subsection{Node-level analysis (\ref{q:nodes})}
\label{sec:node-scores}

In order to study residuals associated with a subset $\phi \subset V$ of nodes/sensors, we propose \emph{node score}
\begin{equation}\label{eq:c-phi}
c_\lambda(\phi)\eqdef c_\lambda(\vs_\phi)
\end{equation}
evaluated on subgraph $\vs_\phi$ extracted from multiplex $\vg^*=(V^*,E^*,\vw,\vr)$.
Different from Section~\ref{sec:presence}, node score $c_\lambda(\phi)$ is a correlation score \eqref{eq:az-max-adjusted}, not a statistic \eqref{eq:az-test-statistic}; this particular choice for the assessment function follows from the discussion on the score comparability of Section~\ref{sec:comparability}.

Subgraph $\vs_\phi$ is defined by all edges in $E^*$ with at least one ending node related to $\phi$, \ie,
by all nodes from $V^*$ in set
\begin{equation}
V_\phi^*\eqdef \{v_t: v\in\phi\cap V_t,\ t=1,\dots,T\}
\end{equation}
or the associated 1-hop neighborhood
\begin{equation}
    N(V_\phi^*)\eqdef \bigcup_{v_t\in V_\phi^*} N(v_t);
\end{equation}
$N(v_t)$ denotes the set of 1-hop neighbors of node $v_t$.
Edges in $\vg^*$ linking nodes in $N(V_\phi^*)\setminus V_\phi^*$ are excluded, as they do not directly impact any node of interest $v\in\phi$. 
The special case of analyzing single nodes $v$ derive from \eqref{eq:c-phi} by setting $\phi=\{v\}$, as shown at the central part of Figure~\ref{fig:subset-space}; with little abuse of notation, denote
$c_\lambda(v)\eqdef c_\lambda(\{v\})$.

Relevant considerations follow from inspecting $c_\lambda(v)$. For example, observing large values of $|c_\lambda(v)|$ when relying on global predictive models, like standard graph neural networks sharing parameters across nodes, can be a warning on the presence of local effects that would be better described by local, node-specific models~\citep{montero2021principles,cini2023taming}.

\subsection{Temporal analysis (\ref{q:times})}
\label{sec:time-scores}

Question \ref{q:times} can be addressed by considering subgraphs $\{\vs_t\}_t$ that slice multiplex graph $\vg^*$ along the temporal axis.
For each time step $t$, define \emph{time score}
\begin{equation}
c_\lambda(t)\eqdef c_\lambda(\vs_t)
\end{equation}
determined by the subgraph $\vs_t$ corresponding to the nodes at time steps $t-1$, $t$, and $t+1$.
More specifically, $\vs_t$ is the subgraph of $\vg^*$ defined by those edges with at least one ending node in  
\begin{equation}
V_t^*\eqdef\{v_t:v\in V_t\},
\end{equation}
as depicted at the top of Figure~\ref{fig:subset-time}.

Similarly to $c_\lambda(\phi)$ in \eqref{eq:c-phi}, consider a time window $\omega=\{t_1,t_1+1,\dots,t_2\}$, for some $t_1<t_2$, and construct subgraphs $\vs_\omega$ from
\begin{equation}
    V_\omega^*\eqdef \{v_t:v\in V_t,\ t\in\omega\},
\end{equation}
to define score 
\begin{equation}\label{eq:c-omega}
c_\lambda(\omega)\eqdef c_\lambda(\vs_\omega).
\end{equation}

Discovering time periods associated with large $|c_\lambda(\cdot)|$ might even indicate non-stationary behaviors that the given model is not able to properly capture. 
A strategy to deal with these time-variant processes would consider, \eg, online adaptive mechanisms updating the prediction model over time
\citep{ditzler2015learning}.

\subsection{Local spatio-temporal analysis}
\label{sec:spacetime-scores}

As the number of nodes and time steps increases, scores \eqref{eq:c-phi} and \eqref{eq:c-omega} become less effective in discovering correlations existing in confined regions of the space-time. This limitation arises because $c_{\lambda}(\phi)$ and $c_{\lambda}(\omega)$ are averages across all nodes and all time steps, respectively. 
To address this limitation, we present a strategy for identifying local correlation patterns jointly across time and space.

For each node $v_t\in V^*$ of $\vg^*$ associated with sensor $v$ at time step $t$ define \emph{local score}
\begin{equation}\label{eq:c-tv}
c_\lambda(t,v)\eqdef c_\lambda(\vs_{t,v}),
\end{equation}
as the correlation score evaluated on subgraph $\vs_{t,v}$ constructed from the neighbors of $v_t$; see the bottom part of Figure~\ref{fig:subset-spacetime}.
Correlation score $c_\lambda(t,v)$ is local in the space-time and accounts for the correlation around $(t,v)$.

More intricate correlation patterns can be highlighted by local scores $c_\lambda (t,v)$. An example is provided in Section~\ref{sec:experiment-traffic} where data preprocessing issues have been evidenced.

\subsection{Practical considerations}
\label{sec:practical-considerations}
To conclude the section, we offer further discussion along with remarks and implementation guidance. 

\subsubsection{Assumptions}
\label{remark:assumptions}

The presented AZ-analysis relies on a single main assumption, \ref{a:median}, about the prediction residuals, which ensures that the test statistic $C_\lambda$ and the scoring function $c_\lambda$ have zero expected value in the absence of residual correlation. Moreover, $c_\lambda$ increases as correlation increases, and can be meaningfully compared across subgraphs. 
Assumption~\ref{a:indep}, by contrast, is not required in practice, since the goal of the AZ-analysis is precisely to discover when \ref{a:indep} does not hold; it is used in this paper solely to establish the asymptotic behavior of $C_\lambda$ and $c_\lambda$. 
Finally, \ref{a:weights} is a technical assumption relating to the graph structure that enables the asymptotic analysis of the distribution of $C_\lambda$. In particular, \ref{a:weights} prescribes that the weights are positive -- they are expected to encode the strength of the relations -- and bounded. Both conditions can be enforced by simple transformations that restrict the weights to a desired range. We also recall that weights are optional and, for the sake of this analysis, setting all of them equal to $1$ is a valid alternative.

We comment that violations of Assumption~\ref{a:median} introduce a bias that invalidates the confidence regions of the AZ-whiteness test statistic and makes the correlation scores less interpretable. However, as anticipated in Section~\ref{sec:az-test}, \ref{a:median} is mild. When residuals $\vr_{t,v}$ are scalar values (\ie, $d_y=1$), the equality $\sign(\bar\vr\, \vr_{t,v})=\sign(\bar\vr)\,\sign(\vr_{t,v})$ holds. In this case, the assumption reduces to requiring $\mathbb E_{\vr_{t,v}}[\sign(\vr_{t,v})] = 0$. This is equivalent to $\mathbb P(\vr_{t,v}<0)=\mathbb P(\vr_{t,v}>0)$, that is, the median of $\vr_{t,v}$ is zero. If $m\eqdef \text{median}(\vr_{t,v})\ne 0$, offsetting the residuals by $-m$ resolves the issue. In particular, note that $\vr$ and $\vr-m$ have the same correlation. Moreover, this offset is applied only to assess the presence of correlation and not to measure the prediction accuracy. 
Conversely, in the multivariate case, Assumption \ref{a:median} cannot always be satisfied simply by re-centering the residuals. In such cases, we suggest computing correlation scores $c_\lambda(\cdot)$ (or statistics $C_\lambda(\cdot)$) separately for each of the $d_y$ components. This allows enforcing \ref{a:median} component-wise. The resulting $d_y$ values can then be re-aggregated, \eg, by averaging.

\subsubsection{Source of correlation and model improvement}
\label{remark:guidelines}

While the AZ-analysis highlights where correlation is present in the prediction residuals, it does not by itself identify the underlying cause. 
In practice, correlation may arise from multiple factors, including non-stationarities in the data-generating process, issues in the acquisition or preprocessing pipeline, or patterns that the predictive model fails to capture. 
Sections~\ref{sec:presence}--\ref{sec:spacetime-scores} illustrate examples of how different correlation patterns can be interpreted and addressed.
The AZ-analysis should be viewed as a diagnostic tool that guides the practitioner toward aspects of the system that warrant further investigation. 
Depending on the identified pattern, interventions to improve model quality may involve redesigning the temporal or spatial processing pipelines, revising the training procedure, or adapting how missing and heterogeneous data are handled.

\subsubsection{Choice of the parameters}
\label{remark:hyperparameter}

Setting parameter $\lambda$ to either $0$ or $1$ allows focusing the analysis on temporal or spatial correlation in isolation, respectively. 
The AZ-analysis mainly inspects the scores for $\lambda=0$ and $\lambda=1$, which are complementary, as shown throughout the empirical Sections~\ref{sec:synth},~\ref{sec:experiment-traffic},~and~\ref{sec:experiment-energy}.
Varying $\lambda$ within $(0,1)$ effectively provides an interpolation of the scores $c_0(\cdot)$ and $c_1(\cdot)$ (and similarly for statistics $C_0(\cdot)$ and $C_1(\cdot)$); this might be convenient for visualization purposes, \eg, when reporting local scores $c_\lambda(t,v)$ as in Figures~\ref{fig:az-scores-synth} and \ref{fig:la-gwnet}. 
In particular, $\lambda=1/2$ provides an equal weighting of the temporal component $\widetilde C_{\tim}$ and the spatial one $\widetilde C_{\spa}$ (see \eqref{eq:az-test-statistic} and \eqref{eq:az-max-adjusted}). Unless there is a specific motivation to consider other values, we suggest focusing on $\lambda=0,1/2,1$.

A second parameter of the analysis is the threshold $\gamma$, which is used to identify statistically significant correlations.
According to Theorem~\ref{theo:az-test}, $\gamma$ can be conveniently derived from the quantiles of the standard Gaussian distribution, regardless of the number of edges (and associated weights) of the input graph; therefore, the same threshold can be used for any graph. 
However, we suggest using the AZ-whiteness test mainly in preliminary phases to assess the overall presence of correlation (Section~\ref{sec:presence}), as scanning and performing multiple hypothesis tests for several subgraphs can lead to an overall probability of type-I errors for the comprehensive analysis that differs significantly from the nominal level of each individual test.
In practice, lower thresholds increase the number of detections, while larger thresholds reduce false alarms at the cost of lower sensitivity.

\medskip 
\subsubsection{Handling small subgraphs}
\label{remark:khops}
In sparse graphs, local scores $\{c_\lambda(t,v)\}$ based on 1-hop neighborhoods may lead to high score variance. 
In fact, this phenomenon results from considering subgraphs with very few edges, and can already be observed in Figure~\ref{fig:score-growth-c}, where the standard deviation decreases as the number of edges increases.
In such cases, the subgraph $\vs_{t,v}$ can be constructed from the $k$-hop neighborhood for some $k\ge 2$. This ensures that the number of edges grows approximately as $\delta^k$, where $\delta$ is the average node degree. For instance, in the experiments of this paper, we considered $k=4$ hops around each space-time location $(t,v)$.
The effect of considering different values of $k$ is shown in Figure~\ref{fig:khops} and discussed in Section~\ref{sec:synth} below.

\subsubsection{Computational and memory complexity}
\label{remark:complexity}
The computational and memory complexities of the AZ-whiteness test statistic $C_\lambda(\vs)$ and the correlation scores $c_\lambda(\vs)$ scale linearly in the number of spatio-temporal edges and in the residual dimension, respectively.
Computing either $c_\lambda(\vs)$ or $C_\lambda(\vs)$ for a generic graph $\vs$ and $\lambda\in(0,1)$ has complexity linear in the number of edges $|E_{\vs}|$. 
The complexity of computing $\sign(\vr_{t,u}^\top \vr_{\tau,v})$ is $O(d_y)$, and this operation is carried out for each of the $|E_{\vs}|$ edges. Since the denominators in \eqref{eq:az-test-statistic} and \eqref{eq:az-max-adjusted} are also computed in $O(|E_{\vs}|)$ time, the total complexity is $O(d_y |E_{\vs}|)$.
In a typical spatio-temporal setting with $N$ sensors, $T$ time steps, and average spatial degree $\delta$ (per $\vg_t$), we have approximately $\delta N T$ spatial edges and $N(T-1)$ temporal edges,
so $|E_{\vs}| = O\bigl((\delta+1) N T\bigr)$ and hence a computational complexity of $O\bigl((\delta+1) N T d_y\bigr).$
When $d_y$ and $\delta$ are fixed and moderate, the cost is effectively linear in the number of space-time nodes $|V_{\vs}| \approx N T$.
More specifically, for the correlation scores, the computation is $O(\delta T)$ for the node scores $c_\lambda(v)$, $O(\delta N)$ for the time scores $c_\lambda(t)$, and $O(\delta^k)$ for the local scores $c_\lambda(v,t)$ computed over $k$ hops. 
Regarding memory, the test requires access to the residuals and the edges. Storing all residuals costs $O(|V_{\vs}|\,d_y) = O(N T d_y)$ memory. The edges require $O(|E_{\vs}|)$ memory, which is approximately $\delta N T$ but can be reduced to $O(\delta N)$ when the spatial edges do not vary with time. 
Finally, we note that static graphs enable the use of sparse operations and the parallelization available in modern (graph) deep learning libraries.

\section{Empirical validation of score behavior}
\label{sec:synth}

The first set of experiments considers a synthetic dataset where we artificially induce temporal and spatial correlation to validate the proposed scores $c_\lambda(\cdot)$ and show that they are indeed able to detect correlation patterns. 
After presenting the generation of the synthetic residuals in Section~\ref{sec:synth-data}, the remainder of the section focuses on the AZ-analysis, commenting on each step of the proposed procedure, validating its effectiveness under missing and heterogeneous data, and comparing it with established approaches from the literature. Two real-world use cases follow in Sections~\ref{sec:experiment-traffic} and \ref{sec:experiment-energy}.

\subsection{Synthetic residuals}\label{sec:synth-data}

The residual signal $\vr$ is modeled as Gaussian noise exhibiting spatial and temporal correlations within specific regions defined by sets $A$ and $B$, respectively. Outside $A\cup B$ residuals are \iid.

\begin{figure}
    \centering
    \includegraphics[scale=1]{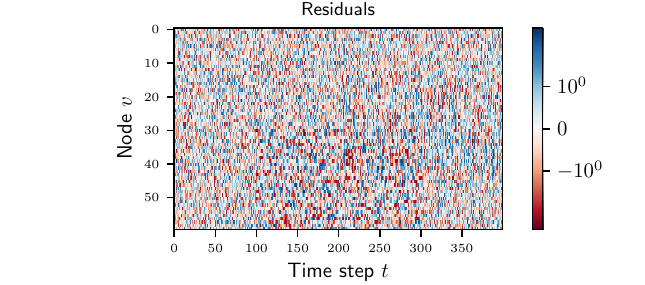}
    \caption{Synthetic residuals generated as described in Section~\ref{sec:synth-data}.}
    \label{fig:synth-data-residuals}
\end{figure}

\begin{figure}
    {\small AZ-whiteness test statistics:
    $\quad C_{0}(\vg^*)=16.2$, 
    $\quad C_{1/2}(\vg^*)=22.9$,
    $\quad C_{1}(\vg^*)=16.1$.}
    \centering
    \includegraphics[width=1\linewidth]{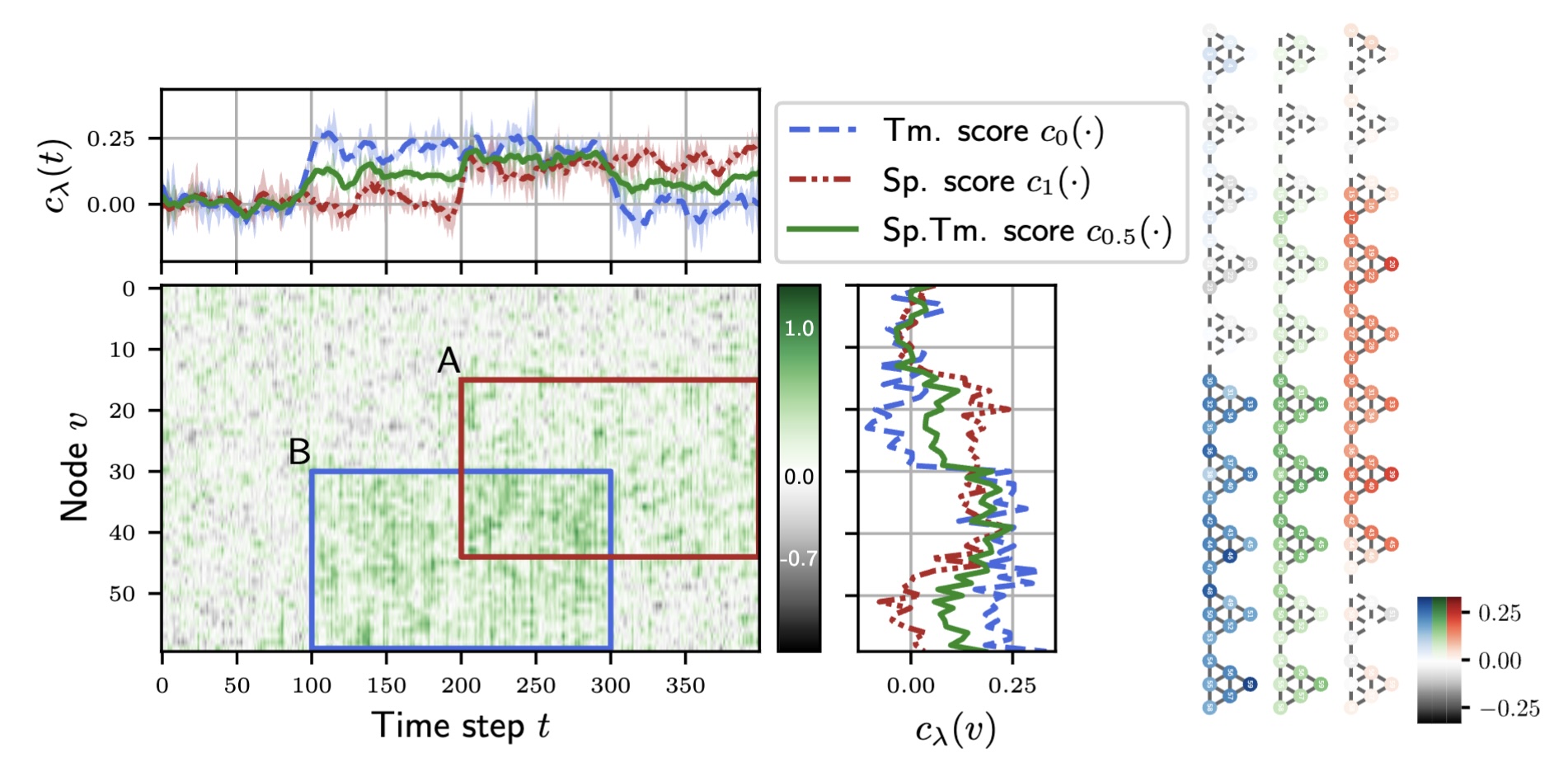}
    \caption{Scores involved in the AZ-analysis of residuals on synthetic data. 
    Scores associated with $\lambda=0$, $1/2$, and $1$ are depicted in blue, green, and red colors, respectively. 
    Top-left) Time scores $c_\lambda(t)$; a moving average is applied to improve readability.
    Center) Node-level scores $c_\lambda(v)$ as both line plots and heatmaps on the graph. 
    Bottom-left) Local spatio-temporal score $c_\lambda(t,v)$ for $\lambda=1/2$; see Figure~\ref{fig:khops} for $\lambda=0$ and $1$.
    Red and blue boxes (sets $A$ and $B$) highlight regions with spatial and temporal correlation, respectively. Values of the AZ-whiteness test statistics are reported at the top left of the figure.
    }
    \label{fig:az-scores-synth}
\end{figure}

The data are generated for 60 nodes and 400 time steps. Set $A$ covers time steps in $[200, 400)$ and nodes in $[15, 45)$, whereas set $B$ covers time steps in $[100, 300)$ and nodes in $[30, 60)$. The considered graph topology and sets $A$ and $B$ are depicted in Figure~\ref{fig:az-scores-synth}.
Starting from \iid white noise $\varepsilon_{t,v}\sim \mathcal N(0,1)$, correlation is induced as follows.
Within region $A \setminus B$, where only spatial correlation is present, residuals are obtained as
\begin{equation}
r_{t,v} = \varepsilon_{t,v} + \frac{\sum_{u\in N(v)} \varepsilon_{t,u}}{|N(v)|}, 
\end{equation}
while in $B \setminus A$ only temporal correlation is present:
\begin{equation}
r_{t,v} = \varepsilon_{t,v} + \frac{\varepsilon_{t-1,v} + \varepsilon_{t+1,v}}{2}.
\end{equation}
Both spatial and temporal correlations are introduced in $A \cap B$:
\begin{equation}
    r_{t,v} = \varepsilon_{t,v} + \frac{\sum_{u\in N(v)} \varepsilon_{t,u} }{|N(v)|}
+ \frac{\varepsilon_{t-1,v} + \varepsilon_{t+1,v}}{2}. 
\end{equation} 
Finally, for each point $(t,v)$ outside $A\cup B$, uncorrelated residuals $r_{t,v}$ equal $\varepsilon_{t,v}$.
An example of correlated residuals generated from this process is depicted in Figure~\ref{fig:synth-data-residuals}.

\begin{figure}
    \includegraphics[width=1\linewidth]{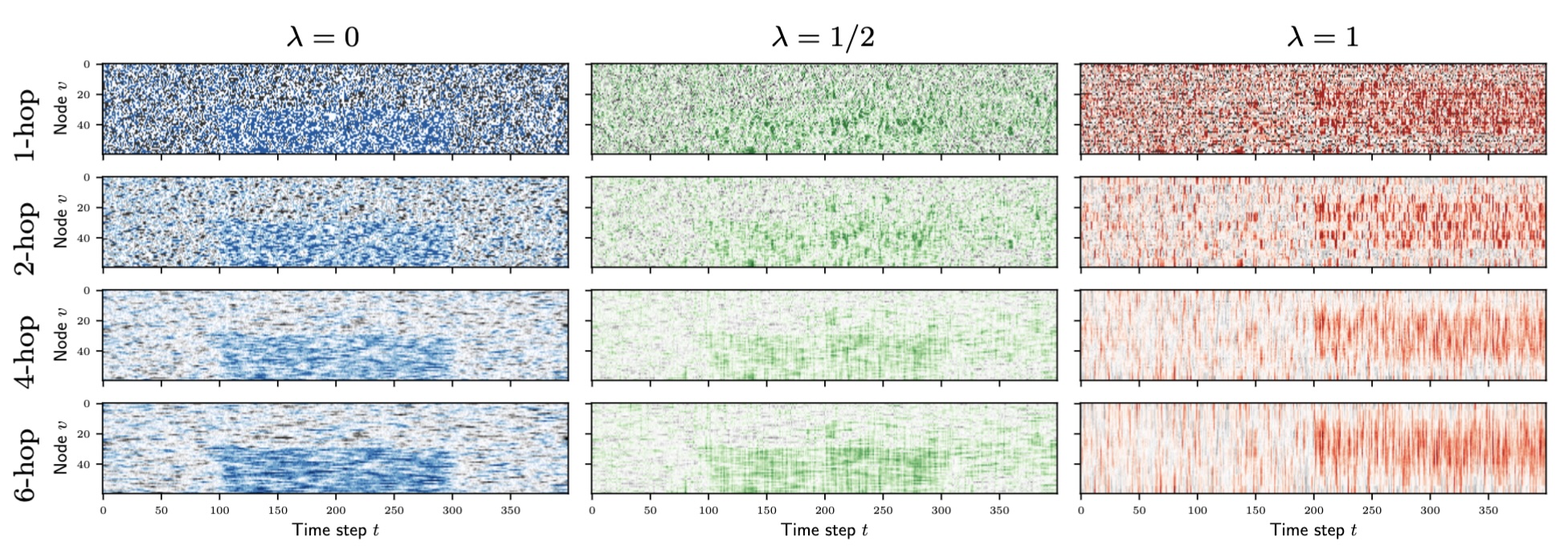}
\caption{Local scores $c_\lambda(t,v)$ computed over different $k$-hop neighborhoods, $k=1, 2, 4,$ and $6$, on the synthetic data described in Section~\ref{sec:synth-data}. $k=4$ is the value used in the other figures, unless stated otherwise. Unlike the other figures in this section, the aspect ratio is set close to 1 to better visualize the smoothing effects along both axes. The colormap is consistent across the plots.}
\label{fig:khops}
\end{figure}

\subsection{Analysis of residuals}\label{sec:exp-synth-base}
Figure~\ref{fig:az-scores-synth} provides a visual representation of the proposed analysis of residuals. We observe the following.
\begin{description}
    \item[Overall presence of correlation] 
    The values of the AZ-whiteness test statistics $C_\lambda(\vg^*)$ are substantially larger\footnote{For reference, setting $\alpha=10^{-5}$ in \eqref{eq:significance-level} leads to $\gamma \approx 4.3$.} than $0$, correctly identifying the overall presence of both spatial and temporal correlation.
    \item[Spatial correlation] 
    The magnitude of all node-level scores $c_{\lambda=1}(v)$ and time scores $c_{\lambda=1}(t)$ (red dot-dashed lines) is consistent with set $A$ containing spatial correlation.
    \item[Temporal correlation] 
    Similarly, $c_{\lambda=0}(v)$ and $c_{\lambda=0}(t)$ (blue dashed lines) are consistent with $B$ being associated with temporal correlation. 
    \item[Spatio-temporal scores] 
    $c_{\lambda=1/2}(v)$ and $c_{\lambda=1/2}(t)$ (green solid lines) are consistent with the union set $A\cup B$; local scores $c_{\lambda=1/2}(t,v)$ have similar behavior. Moreover, the scores are larger where both types of correlation are present ($A\cap B$) than where only one of the two is present ($A\setminus B$ and $B\setminus A$).

\end{description}
The values of the local scores $c_{\lambda}(t,v)$ with $\lambda=1$ and $\lambda=0$ are also consistent with both sets $A$ and $B$; see Figure~\ref{fig:khops}. 
Figure~\ref{fig:khops} also shows the effect of considering $k$-hop neighborhoods with $k>1$ on local scores $c_{\lambda}(t,v)$. In uncorrelated regions, the expected value of $c_\lambda$ is zero. As $k$ increases, these regions become whiter (\ie, closer to zero) because aggregating over more edges reduces the score variance; see also Remark~\ref{remark:khops}. We also observe a smoothing effect due to the larger overlaps between subgraphs.

\begin{figure}
    {\small AZ-whiteness test statistic:
    $\quad C_{0}(\vg^*)=14.7$, $\quad C_{1/2}(\vg^*)=20.8$, $\quad C_{1}(\vg^*)=14.8$.}
    \centering
    \includegraphics[scale=1., trim={0 0 0 2.5cm}, clip]{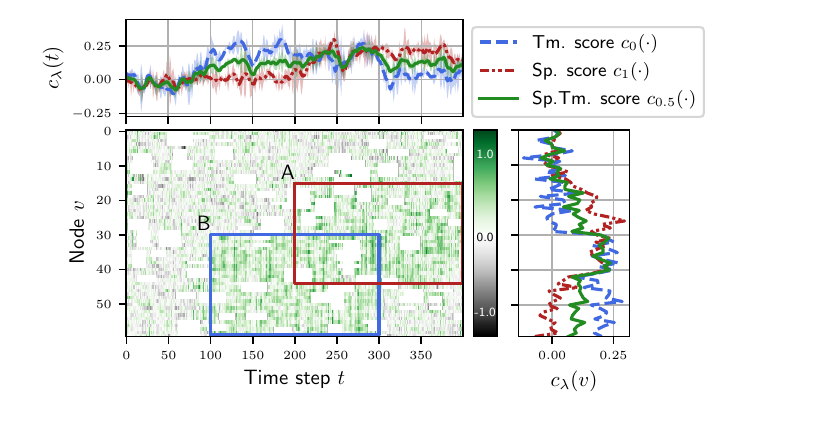}
    \caption{Scores involved in the AZ-analysis of residuals on synthetic residuals with missing observations. White areas in the bottom-left figure represent regions of missing observations.}
    \label{fig:synth-data-missing}
\end{figure}

\begin{figure}
    {\small AZ-whiteness test statistic:  
    $\quad C_{0}(\vg^*)=12.0$, $\quad C_{1/2}(\vg^*)=22.4$, $\quad C_{1}(\vg^*)=14.7$.}
    \centering
    \includegraphics[scale=1., trim={0 0 0 2.5cm}, clip]{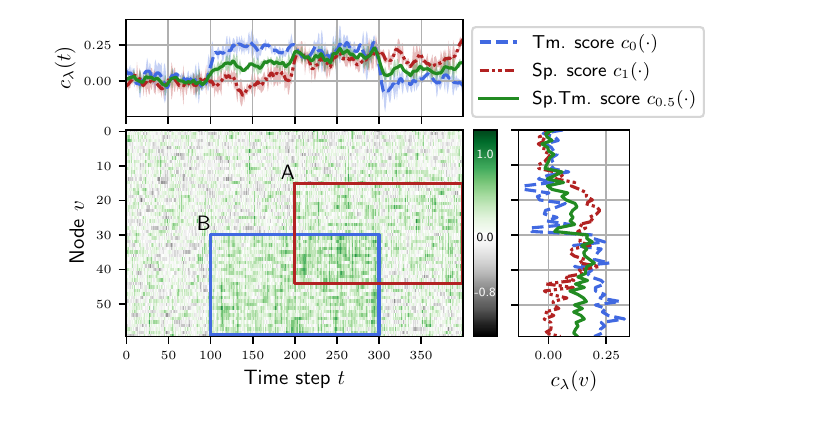}
    \caption{Scores involved in the AZ-analysis of residuals on synthetic residuals from heterogeneous distributions. Residuals at different space-time locations are generated from different distributions.}
    \label{fig:synth-data-heterogeneous}
\end{figure}

\begin{figure}
\centering
    \includegraphics[width=\textwidth]{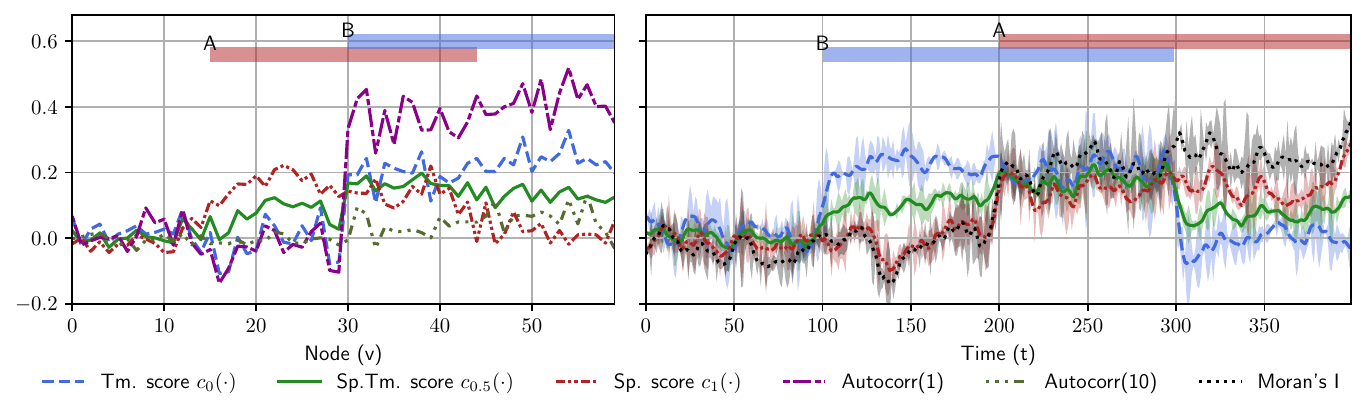}\\
    \includegraphics[trim={6.8cm 0 2.5cm 0}, clip, width=.7\textwidth]{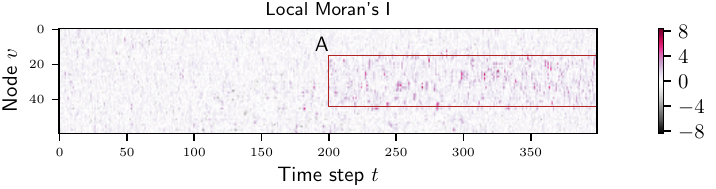}\\
    \includegraphics[trim={6.8cm 0.cm 4.5cm 0.cm}, clip, scale=.66]{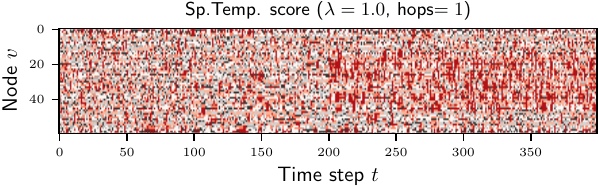}
    \includegraphics[trim={7.6cm 0.cm 2.5cm 0.cm}, clip, scale=.66]{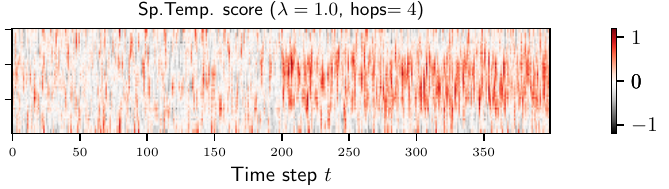}
    \caption{
    Comparison of the proposed correlation scores with established spatial and temporal correlation measures. 
    Top left) node scores $c_\lambda(v)$ and node-wise autocorrelation. 
    Top right) time scores $c_\lambda(t)$ and the time-wise Moran's $I$ statistic. 
    Middle) Local Moran's $I$.
    Bottom) local scores $c_1(t,v)$ computed over 1-hop and 4-hop neighborhoods.}
\label{fig:autocorr-moran}
\end{figure}

\subsection{Analysis with missing and heterogeneous data}
Lastly, we report the scores of the residual analysis carried out on similar synthetic data in which about 20\% of the observations are missing and the data come from heterogeneous sensors. For the scores of these experiments, we refer the reader to Figures~\ref{fig:synth-data-missing} and \ref{fig:synth-data-heterogeneous}.
\begin{description}
    \item[Analysis with missing data] 
    Despite the presence of data gaps, visually represented as white areas within the representation of $c_{\lambda=1/2}(t,v)$ in Figure \ref{fig:synth-data-missing}, 
    the reported correlation scores exhibit patterns aligned with those present in the data and discussed in the previous Section \ref{sec:exp-synth-base}.
\end{description}
To validate the effectiveness of the proposed analysis on data coming from heterogeneous sensors, the residuals $\vr$ are generated following the process described in Section~\ref{sec:synth-data}, but starting from non-\iid\ noise $\varepsilon$. In particular, each noise component $\varepsilon_{t,v}$ is sampled from one of three zero-median distributions: a uniform distribution, a Laplace distribution, a bimodal distribution generated from a mixture of two Gaussian distributions; the distribution assignment is uniform at random.
\begin{description}
    \item[Analysis with heterogeneous data] 
    The analysis reported in Figure~\ref{fig:synth-data-heterogeneous} shows scores very similar to those of Figures \ref{fig:az-scores-synth} and \ref{fig:synth-data-missing}. In particular, both regions $A$ and $B$ are clearly identifiable and analogous conclusions can be drawn.  
\end{description}
We now examine how our scores compare to alternative correlation measures commonly used in temporal and spatial analysis, which provide natural points of reference for assessing the behavior of the proposed scores: $k$-lag autocorrelation and Moran's $I$ statistics~\citep{moran1950notes, anselin1995local}. 
For each node $v$, we consider the aggregated autocorrelation measure
\begin{equation}
\text{Autocorr}_v(\bar k) 
\eqdef 
\frac{1}{\bar k}\sum_{k=1}^{\bar k} \rho_k(v),
\end{equation}
where $\rho_k(v)$ estimates the autocorrelation $\text{corr}(r_{v,t},r_{v,t- k})$ at lag $k>0$.
This provides a summary of temporal dependence up to lag $\bar k$. 
For spatial dependence, we consider Moran's $I$ statistic computed at each time step $t$
\begin{equation}
I_t
\eqdef
\frac{
\sum_{(u,v)\in E_t} w_{t,(u,v)}\,
(r_{t,u}-\bar r_t)\,(r_{t,v}-\bar r_t)
}{
\lVert \vw_t\rVert_1 \; s_t^2
},
\end{equation}
and the Local Moran's $I$ statistic to capture local spatial patterns
\begin{equation}
I_{t,v}
\eqdef
\frac{
\sum_{u\in N_t(v)} w_{t,(u,v)}\,
(r_{t,u}-\bar r_t)\,(r_{t,v}-\bar r_t)
}{
\rVert \{w_{t,(u,v)}: u\in N_t(v)\}\rVert_1\; s_t^2
},
\end{equation}
where $\bar r_t$ denotes the mean residual at time $t$ and $s_t^2 = |V_t|^{-1} \sum_{v\in V_t} (r_{t,v}-\bar r_t)^2$.
\begin{description}
\item[Comparison with other correlation analyses] 
Figure~\ref{fig:autocorr-moran} shows that $c_0(v)$ behaves similarly to Autocorr($k$): both increase in regions with temporal correlation. Autocorr(10) takes lower values than Autocorr(1) because the data exhibit only 1-lag dependence. 
For spatial dependence, the time scores $c_1(t)$ closely match the behavior of Moran's $I$ and identify the same interval of spatial correlation. 
At the local level, both the proposed local scores $c_\lambda(t,v)$ and the Local Moran's $I$ detect region $A$ as spatially correlated, with $c_\lambda(t,v)$ computed over 4-hop neighborhoods doing so more distinctly.
\end{description}
To conclude, the reported results validate the proposed AZ-analysis by showing that both spatial and temporal correlations can be identified. In particular, they can be identified at a global level \eqref{q:global}, at the level of single nodes \eqref{q:nodes}, and at the level of time steps \eqref{q:times}.
Moreover, the AZ-analysis identifies patterns aligned with existing approaches designed for either spatial or temporal correlation, while enabling a comprehensive spatio-temporal analysis of residual correlation under mild assumptions. The results also confirm the broad applicability of the proposed residual analysis, which can effectively operate under mild assumptions and with missing and heterogeneous data.

\section{Use case in traffic forecasting}
\label{sec:experiment-traffic}

In the second experiment, we analyze the prediction residuals of spatio-temporal graph neural networks (STGNNs) trained for multi-step ahead forecasting on traffic flow data.

\subsection{Data and models}

The study utilizes the MetrLA traffic dataset~\citep{li2018diffusion}, which aggregates traffic information into 5-minute intervals of sensor readings from March to June 2012. The dataset comprises 207 univariate time series. A graph connecting the sensors is constructed from their pairwise distance. The provided data contain missing observations (approximately 8\% of the data), which were imputed using preceding observations.

The selected set of predictive models includes generic baselines for time series forecasting and well-established methods from the literature: GWNet~\citep{wu2019graph}, DCRNN~\citep{li2018diffusion}, and AGCRN~\citep{bai2020adaptive}. The baseline models consist of a recurrent neural network (RNN) with gated recurrent units (GRUs) trained and operated on individual univariate time series (uvRNN), a second RNN trained and operated on the multivariate time series obtained by stacking all univariate time series (mvRNN), and a generic time-then-space STGNN (ttsRNN) processing the output of uvRNN with a graph neural network~\citep{cini2023taming}.
Additional experimental details are provided in Appendices~\ref{a:train} and \ref{a:hw-sw}.

\begin{figure}
    \centering
    \begin{subfigure}[t]{.48\textwidth}
        \caption{Multi-step ahead forecasting}
        \label{fig:az-all-temporal-scores-metrla-data}
    \end{subfigure}
    \begin{subfigure}[t]{.48\textwidth}
        \caption{1-step ahead forecasting}
        \label{fig:metrla-1d}
        \vspace{-4em}
    \end{subfigure}
    \includegraphics[width=\textwidth]{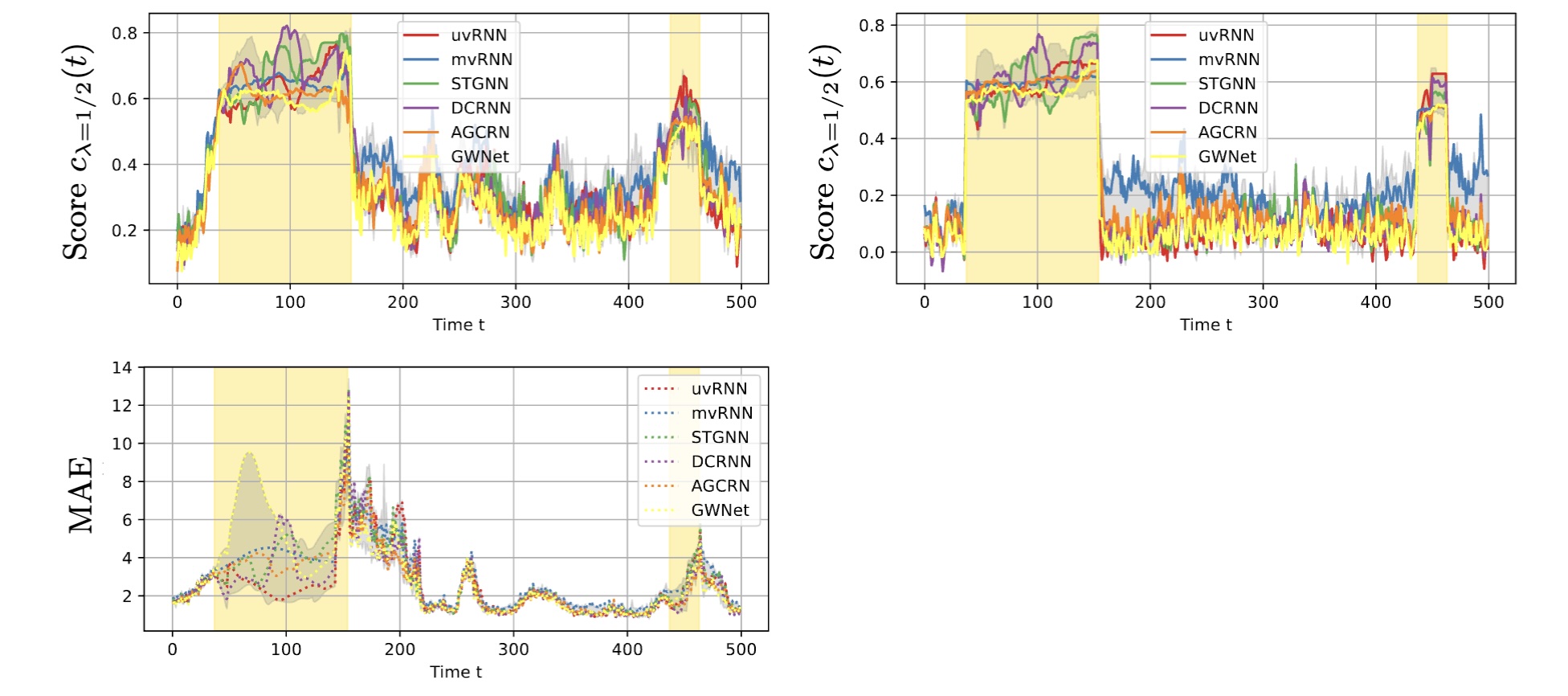}
    \caption{Left) Time scores $c_{\lambda=1/2}(t)$ of the residuals (top) and predictive MAE (bottom) on MetrLA. 
    Right) Time scores $c_{\lambda=1/2}(t)$ on 1-step ahead prediction on MetrLA. 
    The shaded area in gray represents the min-max range of scores/MAE obtained from 18 different models (3 runs for each listed model).
    Time frames highlighted in yellow denote imputed target data.}
    \label{fig:metrla-time}
\end{figure}

\begin{figure}
    \centering
    \includegraphics[width=\textwidth]{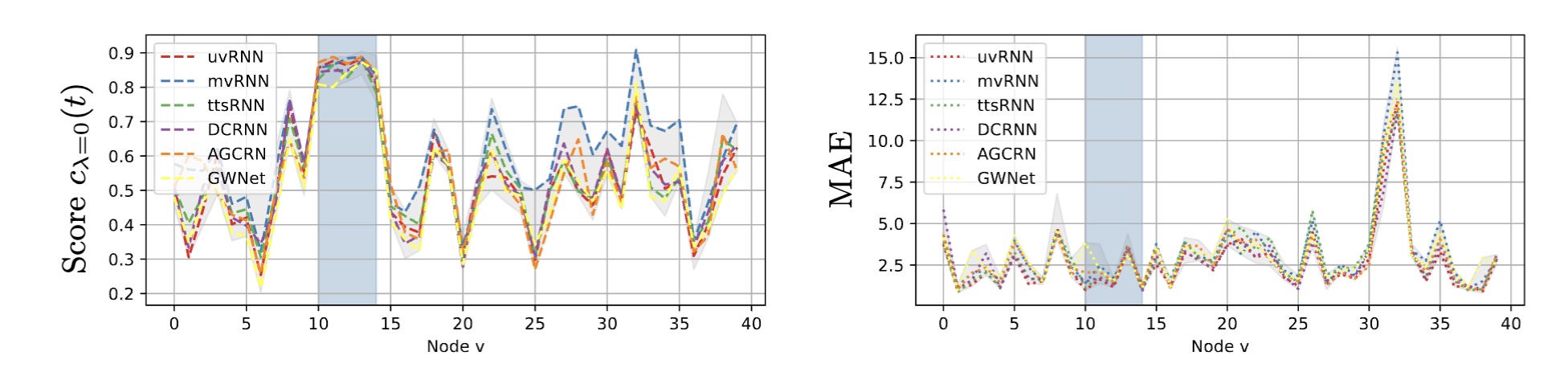}
    \caption{Node scores $c_{\lambda=0}(v)$ of the residuals (left) and predictive MAE (right) on MetrLA.  
    The gray shaded band shows the min--max range across 18 model instances (three runs for each listed model). 
    Nodes highlighted in blue are those where temporal correlation was artificially injected at test time.
    }
    \label{fig:metrla-node-correlation}
\end{figure}

\subsection{Collected insights and discussion} 

Figure~\ref{fig:metrla-time} presents the time scores $c_{\lambda=1/2}(t)$ and prediction error (expressed here in terms of MAE) for all considered models.
For visualization purposes, a window of 500 time steps and 40 nodes is depicted in the figures.

\begin{description}
\item[Identification of correlation patterns]
From the top part of Figure~\ref{fig:az-all-temporal-scores-metrla-data}, we observe that regardless of the predictive model, the time scores are all consistently high within two time intervals around $t=100$ and $t=450$; as indicated by the shaded areas in the figure, these regions correspond to time steps with artificially imputed data. The observed high scores likely relate to the imputation method employed, which replicates the last observed value.

\item[Supplementing performance-based analyses]
The bottom part of Figure~\ref{fig:az-all-temporal-scores-metrla-data} displays the time-wise MAE of all predictors. Comparing the MAE with scores $c_{\lambda=1/2}(t)$, we observe that the higher correlation in the shaded areas is not accompanied by a significant increase in prediction error. This finding demonstrates that the proposed residual analysis can reveal valuable insights that are not apparent from analyses solely based on prediction error

\item[Predictive horizon] 
Comparing Figure~\ref{fig:az-all-temporal-scores-metrla-data}, which computes time scores from multi-step ahead predictions, with Figure~\ref{fig:metrla-1d} focusing on 1-step ahead predictions, reveals that the correlation patterns are more pronounced in the latter case. Furthermore, the observed correlation is diminished outside the imputed regions, suggesting that there is less room for improvement in short-term predictions compared to long-term predictions.

\end{description}

To corroborate the ability of the proposed scores to identify correlation, we manually introduced temporal dependence in the test data for five nodes highlighted in blue in Figure~\ref{fig:metrla-node-correlation}. 
Specifically, we enforced correlation by applying a moving average of width three: $\widetilde \vx_{t,v} = (\vx_{t-1,v} + \vx_{t,v} + \vx_{t+1,v})/3$.
Figure~\ref{fig:metrla-node-correlation} reports the resulting node scores and prediction MAE across all considered models. 
The injected correlation is clearly reflected by the proposed node scores $c_{\lambda=0}(v)$, whereas the MAE appears essentially unaffected.

\begin{figure}
    \includegraphics[width=\textwidth]{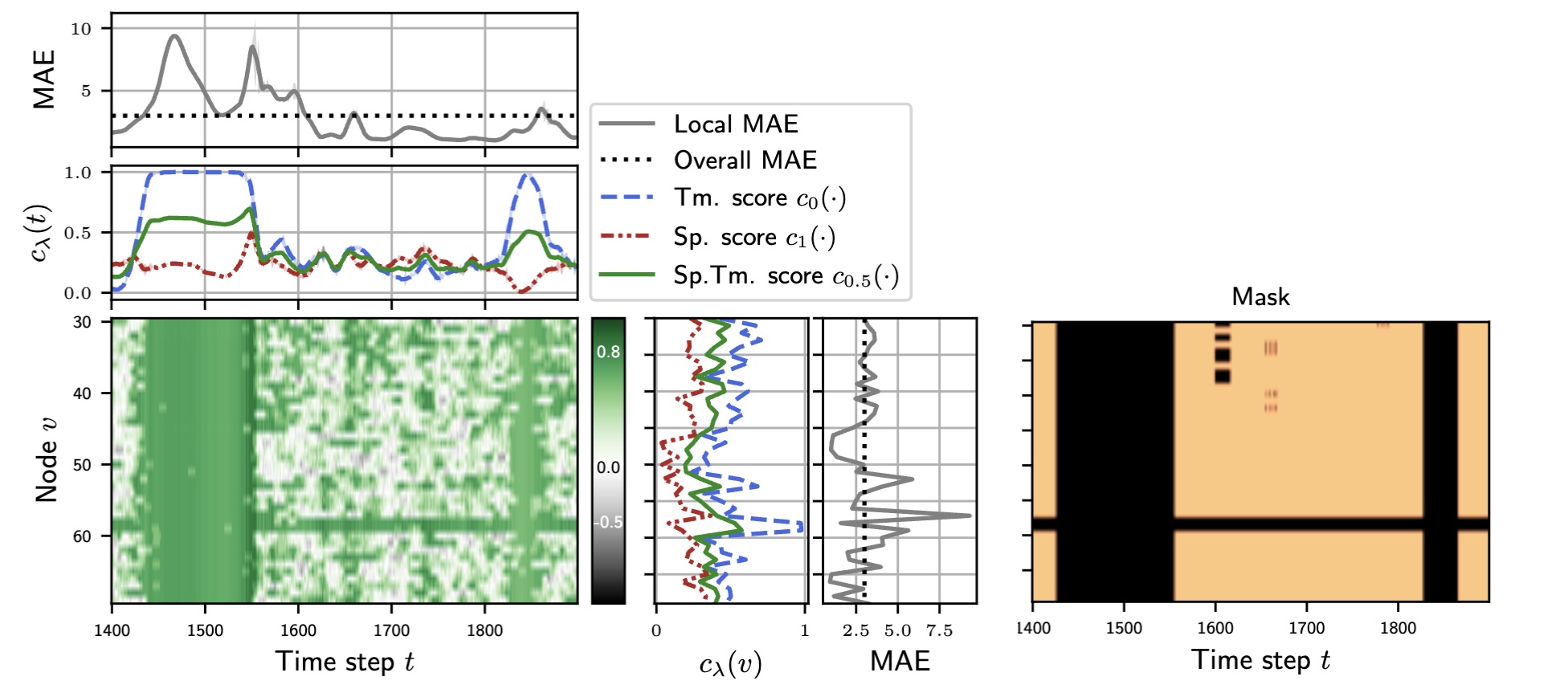}
    \caption{Extended residual analysis performed on MetrLA for a single model. Regions of the imputed data are reported as dark regions in the plot on the right. 
    }
    \label{fig:la-gwnet}
\end{figure}

Extending the residual analysis for one of the models, GWNet, yields further relevant insights; refer to Figure~\ref{fig:la-gwnet}. 
\begin{description}
\item[Local correlation patterns]
Scores $c_\lambda(t,v)$ confirm that the aforementioned time frames are indeed problematic. Moreover, they highlight specific nodes, namely nodes 28 and 29, that warrant closer inspection. Further examination of the mask of imputed data at the bottom of Figure~\ref{fig:la-gwnet} reveals that the time series for these nodes have also been imputed.
\end{description}

\begin{figure}
    \centering
    \includegraphics[width=.98\textwidth]{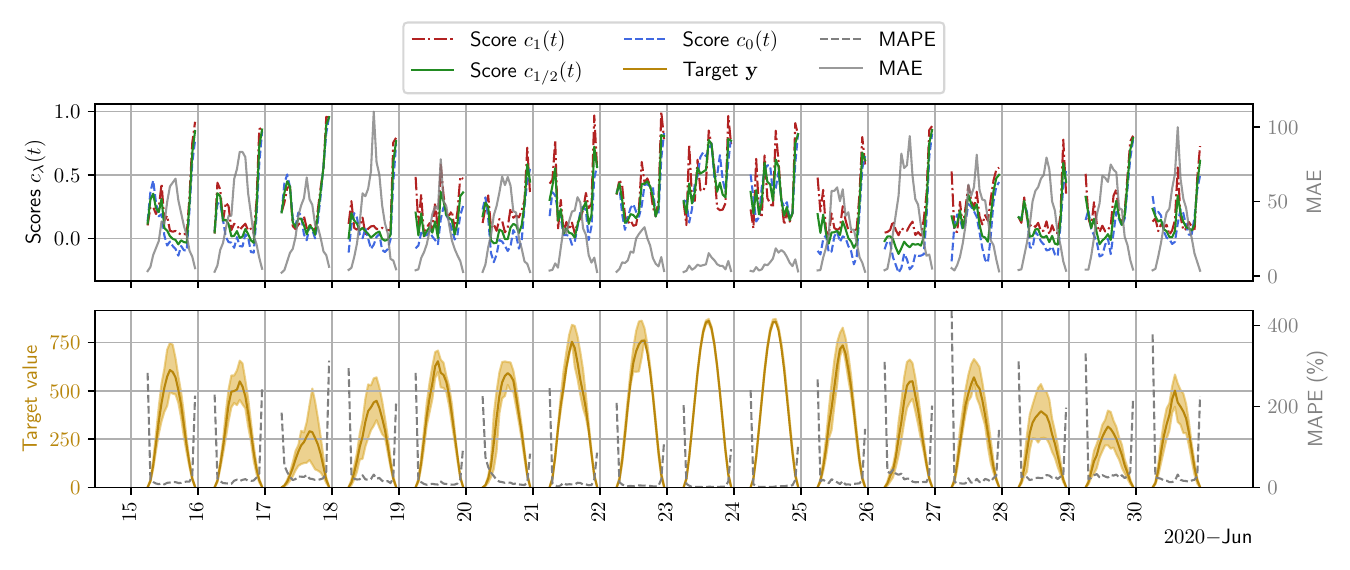}
    \caption{Analysis of the prediction residuals for energy production prediction. 
    Top) Time scores $c_{\lambda}(t)$ and MAE of three days of energy production. 
    Bottom) Relative prediction error (MAPE) and mean absolute value of the target $\vy_{t}$. Absent observations correspond to the null solar radiation during nighttime. Shaded area denote the interquartile range.}
    \label{fig:engrad}
\end{figure}

\begin{figure}
    \centering
    \scriptsize
    \begin{subfigure}[b]{.49\textwidth}
    \includegraphics[scale=.65, trim={.3cm .1cm 1.4cm 1.6cm}, clip]{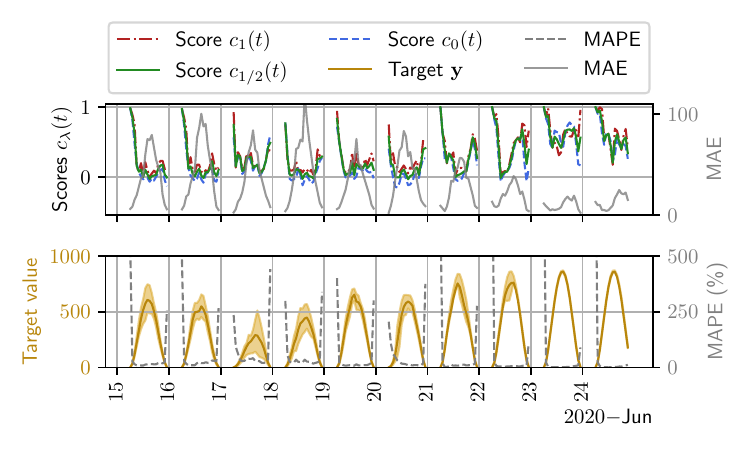}%
    \caption{uvRNN}
    \end{subfigure}
    \begin{subfigure}[b]{.49\textwidth}
    \includegraphics[scale=.65, trim={1.6cm .1cm 0.3cm 1.6cm}, clip]{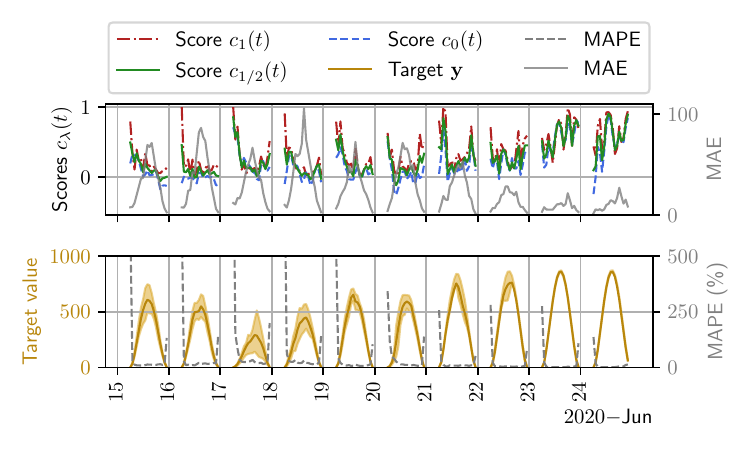}%
    \caption{DCRNN\hspace{2em}~}
    \end{subfigure}
    \bigskip
    
    \begin{subfigure}[b]{.49\textwidth}
    \includegraphics[scale=.65, trim={.3cm .1cm 1.4cm 1.6cm}, clip]{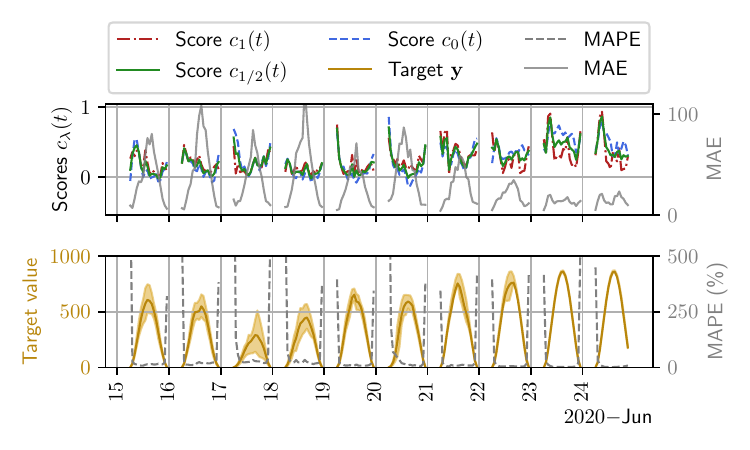}%
    \caption{AGCRN}
    \end{subfigure}
    \begin{subfigure}[b]{.49\textwidth}
    \includegraphics[scale=.65, trim={1.6cm .1cm 0.3cm 1.6cm}, clip]{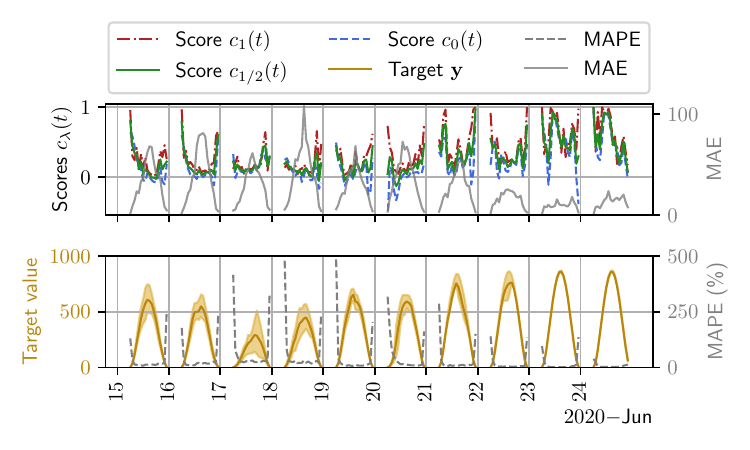}%
    \caption{GWNet\hspace{2em}~}
    \end{subfigure}
    \caption{Analysis of the prediction residuals for energy production prediction, extending Figure~\ref{fig:engrad} to other models.}
    \label{fig:engrad-comparison}
\end{figure}

\section{Use case in energy production}
\label{sec:experiment-energy}

In the third experiment, we investigate the task of forecasting energy production from photovoltaic plants. 

\subsection{Data and models} 
We consider the EngRAD dataset~\citep{marisca2024graph}, which encompasses 5 meteorological variables related to photovoltaic energy production. Hourly estimates of these variables are obtained from a grid of 487 locations across England. A graph representation is constructed based on the pairwise geographical distance between locations. The ttsRNN model was employed and trained to minimize the MAE on 3-hour-ahead forecasts of horizontal irradiance, leveraging an input window of 24 preceding hours and related weather variables. 
The residual analysis was conducted on the test set, focusing on 1-hour-ahead forecasts. Further experimental details are provided in \ref{a:train} and \ref{a:hw-sw}.

\subsection{Insights from the residual analysis} 
Figure~\ref{fig:engrad} reveals a clear daily trend in both the MAE and the time scores $c_\lambda(t)$, reflecting the natural solar irradiation cycle. 
Missing observations correspond to nighttime periods with negligible irradiance.

\begin{description}
    \item[Correlation at dawn and dusk]
    Higher correlations observed during dawn and dusk suggest room for model improvement, while the MAE remains relatively low throughout these transitional periods. While the MAE alone is not particularly concerning during dawn and dusk, further inspection reveals a higher relative prediction error (mean absolute percentage error, MAPE), reinforcing the evidence provided by $c_{\lambda}(t)$ that the model's predictions can improve in these hours of the day.
\end{description}

Beyond these effects, the residual analysis unveils an additional pattern: in certain instances where the model exhibits low MAE values, a high correlation is present, suggesting that the predictions could be further refined.
For example, on June 23rd and 24th, 2020, the target variable attains a notably high magnitude with reduced variability, coupled with a consistently lower MAE. 
While this suggests that the target variable is easier to forecast during these specific periods, it is plausible that the model training is more strongly influenced by regions characterized by higher prediction errors.
Analogous behavior is visible for different models too, as shown in Figure~\ref{fig:engrad-comparison}.

\section{Conclusions}
\label{sec:conclusions}
In this paper, we introduce a novel analysis of model prediction residuals, called AZ-analysis, for spatio-temporal data. The proposed approach allows practitioners to identify regions, such as time intervals or groups of sensors, where model performance can be improved. 
This analysis complements traditional prediction-error evaluations by assessing temporal and spatial correlation in the residuals. It helps reveal aspects of the predictive model that may be suboptimal.
The AZ-analysis is well-suited for assessing deep neural models in real-world scenarios. It offers several advantages: (i) it can operate in the presence of missing and heterogeneous data; (ii) it scales to large and complex datasets by exploiting graph side information; and (iii) it requires only that residuals are centered around zero in a few specific setups. No further assumptions about the residuals' distribution are necessary -- not even identical distributions. 
We first validate the proposed framework through synthetic experiments and provide general guidance on how to interpret the results. We then showcase its potential through real-world use cases in forecasting traffic flow and energy production. 
Overall, AZ-analysis offers a powerful and widely applicable tool for gaining deeper insights into the behavior and limitations of spatio-temporal predictive models.

\appendix

\section{Experiment on the comparability of scores}\label{a:score-growth}

In the experiment of Figure~\ref{fig:score-growth}, Section~\ref{sec:comparability}, the values of statistic $C_\lambda(\vs)$ and score $c_\lambda(\vs)$ are computed for graphs $\vs$ of different number of edges $|E_{\vs}|$ and Pearson correlation among residuals. Considered subgraph $\vs$ is given as a collection of non-incident edges. For each edge $(u,v)$, residual $\vr_v\in\R^d$ of node $v$ is generated as 
\begin{equation}
    \vr_v = \rho \,\vr_u + \sqrt{1-\rho^2} \,\vr_u' + \vm,
\end{equation} 
where $\vr_u$ is the residual at node $u$. The components of vectors $\vr_u$ and $\vr_u'$ are independent samples from the standard Gaussian distribution, whereas $\vm$ centers the median of $\vr_v$ to zero.
Note that scalar $\rho$ is selected in the $[0, 1]$ interval and determines the Pearson correlation between $\vr_v$ and $\vr_u$.
In fact, as $\EE[\vr_u]=\mathbf 0$, $\var[\vr_u]=\mathbb I$ and $\EE[\vr_v] = \vm$ by construction,
\begin{equation}
    \var[\vr_v]=\rho^2 \var[\vr_u] + (1-\rho^2) \var[\vr_u'] = \mathbb I
\end{equation}
and 
\begin{align}
    \text{Cor}[\vr_u,\vr_v] 
    &= \EE[(\vr_v - \vm) \vr_u^\top] = 
  \\&= \EE[( \rho \,\vr_u + \sqrt{1-\rho^2} \,\vr_u') \vr_u^\top]
  \\&= \rho\EE[\vr_u\vr_u^\top] + \sqrt{1-\rho^2} \EE[\vr_u'\vr_u^\top] 
  \\&= \rho\, \mathbb I + \sqrt{1-\rho^2}\EE[\vr_u']^\top \EE[\vr_u] = \rho \,\mathbb I .
\end{align}
Moreover, above construction grants Assumption~\ref{a:median} to hold.

Dimension $d$ of residual vectors is set to 5. 
Mean, standard deviation, and 25th and 75th percentiles are reported in Figure~\ref{fig:score-growth} and are estimated over 100 repeated simulations.

\section{Additional details on models and training}\label{a:train}

For the traffic forecasting task, predictions are generated from an input window of 1 hour (12 time steps) to forecast the subsequent hour. The hour of the day and the day of the week are included as covariates for all models.
For the solar irradiance task, forecasts are issued 6 hours ahead using an input window of 24 hours; in this case, the hour of the day and the hour of the year are included as covariates.
Residual analyses are conducted on the test set considering traffic prediction horizons of 5, 20, 40, and 60 minutes, and 1-hour-ahead solar irradiance forecasts.

For the traffic forecasting task, predictions are made from an input window of 1 hour (12 time steps) to predict the subsequent hour. The hour of the day and the day of the week are included as covariates for all models. 
For the solar irradiance prediction, forecasts are issued 6 hours ahead using an input window of 24 hours; here, the hour of the day and hour of the year are added as covariates.
The residual analyses are conducted on the test set considering traffic forecasts at 5, 20, 40, and 60 minutes, and 1-hour-ahead solar irradiance forecasts. 
The graphs are derived from pairwise geographical distances between sensors, which are processed through a Gaussian kernel to define edge weights. Edges with a weight below 0.1 are discarded; for EngRAD, connections beyond the 8 nearest neighbors are also removed.

Baseline models uvRNN, mvRNN, and ttsRNN are single-layer recurrent neural networks with gated recurrent units and exponential linear unit activation. uvRNN and mvRNN use 32 hidden units. ttsRNN uses 16 hidden units and implements a simple 2-layer message-passing scheme with average aggregation on the uvRNN outputs.
All models are trained to minimize the mean absolute error (MAE), with missing and imputed data appropriately masked during training. 
For MetrLA, the first 70\% of the data is used for training, the next 10\% for validation, and the remaining 20\% for testing. For EngRAD, the first two years are used for training, with 12 weeks across all four seasons reserved for validation, and the final year is used for testing.
Training is performed for 200 epochs using the Adam optimizer with an initial learning rate of 0.003, reduced by a factor of 4 every 50 epochs. Each epoch consists of 300 batches of size 64. Early stopping is applied after 50 epochs without improvement on the validation set.

\section{Hardware, software and data}\label{a:hw-sw}

Model training and inference are performed on a workstation equipped with Intel(R) Xeon(R) Silver 4116 CPU \@ 2.10GHz processors and NVIDIA TITAN V GPUs. The analysis of residuals does not require GPU acceleration.
Experiments are developed in Python 3.8 mainly relying on open-source libraries PyTorch~\citep{paszke2019pytorch}, PyTorch Geometric~\citep{PyTorchGeometric}, Torch Spatiotemporal~\citep{TorchSpatiotemporal} and NumPy~\citep{harris2020array}. 
In particular, Torch Spatiotemporal offers utilities for downloading the data used for the traffic and energy production use cases. Instructions for generating the synthetic residuals are detailed in the main paper.
The source code to reproduce the results is available at: {\small\url{https://github.com/dzambon/az-analysis}}.

\section*{Acknowledgements}
This work was supported by the Swiss National Science Foundation project FNS 204061: \emph{High-Order Relations and Dynamics in Graph Neural Networks}.

\bibliographystyle{abbrvnat}
\bibliography{biblio}

\end{document}